\documentclass{article}

\usepackage[preprint]{nips_2019}
\usepackage[utf8]{inputenc} % allow utf-8 input
\usepackage[T1]{fontenc}    % use 8-bit T1 fonts
\usepackage{hyperref}       % hyperlinks
\usepackage{url}            % simple URL typesetting
\usepackage{amsfonts}       % blackboard math symbols
\usepackage{nicefrac}       % compact symbols for 1/2, etc.
\usepackage{microtype}      % microtypography

\usepackage{amsmath,amssymb, amsthm}

\DeclareMathOperator*{\softmax}{softmax}

\usepackage{graphicx}
\usepackage{subcaption}
\usepackage{adjustbox}

\usepackage{booktabs}
\usepackage{multirow}
\usepackage{algorithm}
\usepackage[noend]{algpseudocode}
\usepackage{natbib}

\title{Confidence Calibration for Convolutional Neural Networks Using Structured Dropout}

\author{Anonymous \\
 \\
}
\author{
 Zhilu Zhang \\
 Cornell University \\
 \texttt{zz452@cornell.edu} \\
 \And
 Adrian V.~Dalca \\
 MIT and MGH\\
  \texttt{adalca@mit.edu} \\
  \And
  Mert R.~Sabuncu  \\
  Cornell Univerisity \\
  \texttt{msabuncu@cornell.edu}
}
\begin{document}

\maketitle

\begin{abstract}
%Obtaining calibrated probabilistic predictions from Neural networks (NN) models can be crucial. 
In classification applications, we often want probabilistic predictions to reflect confidence or uncertainty.
Dropout, a commonly used training technique, has recently been linked to Bayesian inference, yielding an efficient way to quantify uncertainty in neural network models. 
However, as previously demonstrated, confidence estimates computed with a naive implementation of dropout can be poorly calibrated, particularly when using convolutional networks. 
In this paper, through the lens of ensemble learning, we associate calibration error with the correlation between the models sampled with dropout. 
Motivated by this, we explore the use of structured dropout to promote model diversity and improve confidence calibration.
We use the SVHN, CIFAR-10 and CIFAR-100 datasets to empirically compare model diversity and confidence errors obtained using various dropout techniques. 
We also show the merit of structured dropout in a Bayesian active learning application.    
\end{abstract}

\section{Introduction}
Deep neural networks (NNs) are achieving state-of-the-art prediction performance in a wide range of applications. %However, high accuracy is not enough. 
However, in order to be adopted for important decision making in real worlds scenarios like medical diagnosis and autonomous driving, reliable uncertainty (or its opposite, confidence) estimates are also crucial.
Nevertheless, most modern NNs are trained with maximum likelihood produce point estimates, which are often over-confident~\cite{guo2017calibration}.

Bayesian techniques can be used with neural networks to obtain well-calibrated probabilistic predictions~\cite{mackay1992practical, neal2012bayesian}. 
However, Bayesian methods suffer from significant computational challenges.
Thus, recent efforts have been devoted to making Bayesian neural networks more computationally efficient~\cite{blundell2015weight, chen2014stochastic, louizos2017multiplicative, wu2018deterministic}. 
Monte Carlo (MC) dropout \cite{gal2016dropout}, a cheap approximate inference technique which obtains uncertainty by performing dropout~\cite{srivastava2014dropout} at test time, is a popular Bayesian method to obtain uncertainty estimates for NNs. 
%A major advantage of \textit{MC dropout} is that it doesn't require a significant change in implementation and scales well to modern architectures and datasets.

%Indeed, on top of the additional computational cost incurred, most of other existing Bayesian methods also require significant changes in implementation, hence hard to scale to modern architectures and datasets with a reasonable amount of computational resource. As a sharp contrast, dropout variational approximation incurs no additional computational cost at training time, and can be readily applied to any exiting architectures.

Despite improvements over deterministic NNs, it has been noted that \textit{MC dropout} can still produce over-confident predictions~\cite{lakshminarayanan2017simple}. 
In this paper, we propose a simple yet effective solution to obtain better calibrated uncertainty estimates with minimal additional computational burden. 
Inspired by \cite{lakshminarayanan2017simple}, we view \textit{MC dropout} as an ensemble of models sampled with dropout. 
We show that confidence calibration is related to model diversity in an ensemble, and attribute the poor calibration of \textit{MC dropout} to limited model diversity.
To alleviate the problem, we propose to promote model diversity with a structured dropout strategy. 
We empirically verify that structured dropout can yield models with more diversity and better confidence estimates on the SVHN, CIFAR-10 and CIFAR-100 datasets. 
We also compare our method with \textit{deep ensemble} \cite{lakshminarayanan2017simple}, and show several advantages. 
Furthermore, we demonstrate the merit of better uncertainty estimates in a Bayesian active learning experiment~\cite{gal2017deep}. %Models trained with structured dropout achieve better predictive accuracy.

%(DO WE NEED THIS PARAGRAPH?)The rest of the paper is organized as follows. Section 2 discusses related methods. Section 3 presents the proposed approach of using structured dropout to obtain well-calibrated uncertainty estimates. Section 4 presents and discusses experiments and results. Finally, section 5 concludes our paper. 

\section{Related Work}\label{sec_2}
Dropout was first introduced as a stochastic regularization technique for NNs~\cite{srivastava2014dropout}. 
Inspired by the success of dropout, numerous variants have recently been proposed~\cite{wan2013regularization,goodfellow2013maxout,larsson2016fractalnet,gastaldi2017shake}.
%, MaxOut \cite{goodfellow2013maxout}, FractalNet \cite{larsson2016fractalnet} and shake-shake regularization\cite{gastaldi2017shake}. 
Unlike regular dropout, most of these methods drop parts of the NNs in a structured manner. 
For instance, DropBlock \cite{ghiasi2018dropblock} applies dropout to small patches of the feature map in convolutional networks, SpatialDrop \cite{tompson2015efficient} drops out entire channels, while Stochastic Depth Net \cite{huang2016deep} drops out entire ResNet blocks. 
These methods were proposed to boost test time accuracy. 
In this paper, we show that these structured dropout techniques can be successfully applied to obtain better confidence estimates as well.

As we discuss below, dropout can be thought of as performing approximate Bayesian inference~\cite{gal2016dropout,gal2016theoretically, gal2017concrete, gal2017deep, kendall2017uncertainties} and offer estimates of uncertainty.
%However, none of them considered using structured dropout to obtain better uncertainty estimates. 
Many other approximate Bayesian inference techniques have also been proposed for NNs \cite{blundell2015weight, graves2011practical, kingma2015variational, louizos2017multiplicative, wu2018deterministic}. 
However, these methods can demand a sophisticated implementation, are often harder to scale, and can suffer from sub-optimal performance~\cite{blier2018description}. 
Another popular alternative to approximate the intractable posterior is Markov Chain Monte Carlo (MCMC) \cite{neal2012bayesian}. More recently, stochastic gradient versions of MCMC were also proposed to allow scalability \cite{chen2014stochastic, gong2018meta, ma2015complete, welling2011bayesian}. 
Nevertheless, these methods are often computationally expensive, and sensitive to the choice of hyper-parameters. Lastly, there have been efforts to approximate the posterior with Laplace approximation \cite{mackay1992practical, ritter2018scalable}. 
A related approach, the SWA-Gaussian \cite{maddox2019simple} is another technique for Gaussian posterior approximation using the Stochastic Weight Averaging (SWA) algorithm \cite{izmailov2018averaging}. 

There are also non-Bayesian techniques to obtain calibrated confidence estimates. 
For example, temperature scaling \cite{guo2017calibration} was empirically demonstrated to be quite effective in calibrating the predictions of a model.
A related line of work uses an ensemble of several randomly-initialized NNs \cite{lakshminarayanan2017simple}. This method, called \textit{deep ensemble}, requires training and saving multiple NN models.
It has also been demonstrated that an ensemble of snapshots of the trained model at different iterations can help obtain better uncertainty estimates \cite{geifman2018bias}. 
Compared to an explicit ensemble, this approach requires training only one model. 
Nevertheless, models at different iterations must all be saved in order to deploy the algorithm, which can be computationally demanding with very large models.

\section{Uncertainty Estimates with Structured Dropouts}\label{sec_3}
%Here, we present structured dropout from a Bayesian perspective. 
In this section, we first provide a brief review of regular dropout as a Bayesian inference technique. 
Then, we provide a fresh perspective on dropout based on ensemble learning. 
Finally, we motivate the use of structured dropout to obtain better calibrations in confidence.

\subsection{MC Dropout as Ensembles of Dropout Models}
We assume a dataset $\mathcal{D} = (\boldsymbol{X}, \boldsymbol{Y}) = \{ (\boldsymbol{x}_i, y_i) \}_{i = 1}^n$, where each $ (\boldsymbol{x}_i, y_i) \in (\mathcal{X} \times \mathcal{Y}) $ is \textit{i.i.d.} 
We consider the problem of k-class classification, and let $ \mathcal{X} \subseteq \mathbb{R}^d$ be the input space and $\mathcal{Y} = \{ 1, \cdots , k \}$ be the label space\footnote{Extension to regression tasks is straightforward but left out of this paper.}. 
%A classifier is a function that maps input features to labels $f: \mathcal{X} \rightarrow \mathbb{R}^k $. 
We restrict our attention to NN functions $f_{\boldsymbol{w}}(\boldsymbol{x}): \mathcal{X} \rightarrow \mathbb{R}^k$, where $\boldsymbol{w} = \{ W_i \}_{i = 1}^L$ corresponds to the parameters of a network with L-layers, and $W_i$ corresponds to the weight matrix in the i-th layer. 
We define a likelihood model $p(y|\boldsymbol{x}, \boldsymbol{w} ) = \softmax(f_{\boldsymbol{w}}(\boldsymbol{x}))$.
Maximum likelihood estimation can be performed to compute point estimates for $\boldsymbol{w}$. 

Recently, Gal and Ghahramani \cite{gal2016dropout} proposed a novel viewpoint of dropout as approximate Bayesian inference. 
This perspective offers a simple way to marginalize out model weights at test time to obtain better calibrated predictions, which is referred to as \textit{MC dropout}:
\begin{align}
\label{eqn:mcdropout}
p(y = c | \boldsymbol{x}, \mathcal{D_{\text{train}}}) = \int p(y=c|\boldsymbol{x}, \boldsymbol{w})p(\boldsymbol{w | \mathcal{D_{\text{train}}}})  d \boldsymbol{w} \approx \frac{1}{T} \sum _{t = 1}^T  p(y | \boldsymbol{x}, \boldsymbol{w}^{(t)}),
\end{align}
where $\boldsymbol{w}^{(t)} \sim q(\boldsymbol{w} | \mathcal{D_{\text{train}}})$ is assumed to be independently drawn layer-wise weight matrices: $W_i^{(t)}  \sim \hat{W_i} \cdot \text{diag}(\text{Bernoulli}(p))$, $\hat{W_i}$ is the parameter matrix learned during training, and $p$ is the dropout rate.

In this paper, we view each dropout sample $\boldsymbol{w}^{(t)}$ in Equation \ref{eqn:mcdropout} corresponding to an individual model in an ensemble, where \textit{MC dropout} is performing (approximate Bayesian) ensemble averaging. 
Interestingly, in the original dropout paper, the authors interpreted dropout as an extreme form of model combination with extensive parameter sharing \cite{srivastava2014dropout}. 
The ensemble learning perspective can suggest potential ways to improve the calibration of the confidence estimates in \textit{MC dropout}. 

First proposed by Krogh and Vedelsby \cite{krogh1995neural}, the error-ambiguity decomposition enables one to quantify the performance of ensembles with respect to individual models. 
Let $\{ h_t \} _{t =1}^T$ be an ensemble of classifiers, and $H(\boldsymbol{x}) = \sum_t h_t(\boldsymbol{x}) / T$ correspond to ensemble averaging. 
Note $h_t(\boldsymbol{x}) =  p(y | \boldsymbol{x}, \boldsymbol{w}^{(t)})$ in \textit{MC dropout}.
For simplicity, let's assume a binary classification problem so that $\mathcal{Y} = \{ 0,1 \}$, and that $h_t(\boldsymbol{x}) =  p(y = 1 | \boldsymbol{x}, \boldsymbol{w}_t)$ is a scalar\footnote{The results can be easily generalized to multi-class classification.}. Model ambiguity can be defined as:
\begin{align}
    \alpha(h_t | \boldsymbol{x}) = (h_t(\boldsymbol{x}) - H(\boldsymbol{x}))^2. \nonumber
\end{align}
%Hence, ambiguity can be used to measure model diversity in an ensemble. 
%If one uses the mean squared error $\text{MSE}(h_t | \boldsymbol{x}) = (y - h_t(\boldsymbol{x}))^2 $ to quantify model performance, 
It can be shown that mean squared error, $\text{MSE}(h_t | \boldsymbol{x}) = (y - h_t(\boldsymbol{x}))^2 $, a measure of performance, can be decomposed into:
\begin{align}
\label{eqn:decomposition}
\text{MSE}(H) = \mathbb{E}_{\boldsymbol{x}}[\text{MSE}(H|\boldsymbol{x})] =  \mathbb{E}_{\boldsymbol{x}}[\overline{\text{MSE}}(h | \boldsymbol{x})] - \mathbb{E}_{\boldsymbol{x}}[\overline{\alpha}(h | \boldsymbol{x})],
\end{align}
where 
$$\overline{\text{MSE}}(h | \boldsymbol{x}) = \frac{1}{T} \sum_{t}^T \text{MSE}(h_t | \boldsymbol{x}), 
\textrm{ and } 
\overline{\alpha}(h | \boldsymbol{x}) = \frac{1}{T} \sum_{t}^T \alpha(h_t | \boldsymbol{x}) $$ 
corresponds to the average MSE and ambiguity over individual models. 
Note that average ambiguity is a measure of model diversity in an ensemble.
Equation \ref{eqn:decomposition} suggests that the more accurate and the more diverse the models, the more accurate the ensemble. 
We use MSE instead of the popular negative log likelihood (NLL) due to mathematical convenience. 
The two metrics are closely related, and insights obtained from MSE carry over to NLL.

The quality of confidence estimates is often quantified via expected calibration error: 
\begin{align}
    \textrm{ECE}(H) = \mathbb{E}_{\boldsymbol{x}}[(\mathbb{E}_y[y|H(\boldsymbol{x})] - H(\boldsymbol{x}))^2],
\end{align}
which measures the expected difference between the true class probability and the confidence of the model $H$ \cite{guo2017calibration, kuleshov2015calibrated}. Decomposing MSE in a different manner enables one to relate MSE with ECE:
\begin{align}
\label{eqn:decomposition2}
\text{MSE}(H) &= \mathbb{E}_{\boldsymbol{x}}[(y - H(\boldsymbol{x}))^2] = \mathbb{E}_{\boldsymbol{x}}[(y - \mathbb{E}_y[y|H(\boldsymbol{x})] )^2] + \textrm{ECE}(H),
\end{align}
where $\mathbb{E}_y[y|H(\boldsymbol{x})]$ corresponds to the true probability of $y = 1$ conditioned on $H(\boldsymbol{x})$.

Combining Equations~\ref{eqn:decomposition} and ~\ref{eqn:decomposition2}, we can write ECE as:
\begin{align}
\label{eqn:ECE}
    \text{ECE}(H) = \mathbb{E}_{\boldsymbol{x}}[\overline{\text{MSE}}(h | \boldsymbol{x})] - \mathbb{E}_{\boldsymbol{x}}[\overline{\alpha}(h | \boldsymbol{x})] + \text{Var}_{\boldsymbol{x}}[\mathbb{E}_y[y|H(\boldsymbol{x})]] - \text{Var}_{\boldsymbol{x}}[y].
\end{align}
ECE is therefore dependent on average MSE and model diversity; the more diverse and accurate the individual models, the smaller the ECE.
%and the second term is a \textit{sharpness} term that measures how much variation there is in the true probability across predictions. The sharpness is often very large for modern neural networks because the term is maximized when predictions are close to 0 or 1, and modern NNs produce overly confident predictions \cite{guo2017calibration}.
$\text{Var}_{\boldsymbol{x}}[\mathbb{E}_y[y|H(\boldsymbol{x})]]$ measures the variation of the true class probabilities across the level-sets of the ensemble model $H$~\cite{kuleshov2015calibrated}. Thus for this metric, the numeric values of $H(\boldsymbol{x})$ are not important. It is minimized if $H(\boldsymbol{x})$ is a constant and maximized when $H(\boldsymbol{x}) = f(y)$, for any bijective function $f$. 
One can therefore view $\text{Var}_{\boldsymbol{x}}[\mathbb{E}_y[y|H(\boldsymbol{x})]]$ as a weak metric of accuracy that is not sensitive to calibration. Note $\text{Var}_{\boldsymbol{x}}[y]$ does not depend on the models.
Equation \ref{eqn:ECE} offers an interesting insight. If we hold the average accuracy across individual models constant, one can reduce calibration error by increasing the diversity among the models.

%We can either increase the model ambiguity or decrease the model MSE and sharpness. For instance, compared to standards NNs, temperature scaling \cite{guo2017calibration} achieves better calibration through lowering the MSE and sharpness of prediction using a single parameter tuned on the validation set. In this case, the ambiguity term is zero since only one network is used. \textit{Deep ensemble} attains the objective through utilizing multiple models for prediction, which "increases" model diversity. Ensemble averaging also leads to less sharp predictions due to discrepancies among predictions. 
%While ECE can also be lowered through reducing MSE, thanks to the large efforts focused on improving accuracy, we believe for many well-studied applications this would be a more challenging objective. 
%it is a much more difficult task than the others in general, because models have been tuned to explicitly minimize loss functions like MSE and NLL, while the other aspects have received less attention.

\subsection{Improved Confidence Calibration with Structured Dropout}\label{structure}
The discussion of the previous section, together with recent reports on the lackluster performance of dropout as a regularizer in convolutional NNs \cite{ghiasi2018dropblock}, provides an explanation of the improved quality of predictions of \textit{deep ensembles} over \textit{MC dropout}. 
%For both cases, we assume the ensembles can achieve 
Since image features are often highly correlated spatially, even with dropout, information about the input can still be propagated to subsequent layers.
Thus there is a lack of diversity in models sampled from \textit{MC dropout}, as the different models learn very similar representations during training due to the information leakage.
On the other hand, through random initialization and stochastic optimization, NNs in \textit{deep ensembles} can learn divergent representations \cite{li2015convergent, wang2018towards} and hence exhibit more model diversity, leading to more calibrated uncertainty estimates. 
Moreover, although the individual models in \textit{deep ensembles} can have lower MSE than that of \textit{MC dropout} due to larger effective model complexity, as we will corroborate empirically in Section \ref{diversity}, the contribution of this does not seem to be significant.

Inspired by the above observation, we propose to use structured dropout in lieu of regular dropout to obtain better confidence calibration. In this paper, we consider structured dropout with CNNs at different scales. To be more specific, we compare dropout at the \textit{patch-level} which randomly drops out small patches of feature maps~\cite{ghiasi2018dropblock}, the \textit{channel-level} which drops out entire channels of feature maps at random~\cite{tompson2015efficient}, and \textit{layer-level} which drops out entire layers of CNNs at random~\cite{huang2016deep} . For the rest of the paper, we denote these as \textit{dropBlock}, \textit{dropChannel} and \textit{dropLayer} respectively. We also explicitly refer the original dropout as \textit{dropout}. Following the convention, we denote the test-time sampling of models trained with the aforementioned structured dropout methods as \textit{MC dropBlock}, \textit{MC dropChannel} and \textit{MC dropLayer}. 

Structured dropout promotes model diversity. Through discarding correlated features, structured dropout produces sampled models with more ambiguity by forcing different parts of the networks to learn different representations. 
%Throwing out information in a contiguous region could also reduce the sharpness of predictions, as the networks must make predictions with less information, thereby further reducing the calibration error.
Mathematically, using structured dropout in lieu of regular dropout amounts to only a change of the approximate distribution $q(\boldsymbol{w}|\mathcal{D_{\text{train}}})$ in Equation~\ref{eqn:mcdropout}, so that we are performing Bayesian variational inference with a different class of approximate distributions. 
For instance, in the channel-level dropout, instead of sampling a Bernoulli random variable for each individual element in the weight matrix across all channels, we sample one Bernoulli random variable for each channel. 

While patch- and the channel-level dropout can be easily implemented in a wide variety of convolutional architectures, layer-level dropout requires skip connections so that there is still an information flow through the network after dropping out an entire layer. Thankfully skip connections are quite popular in modern NNs. Some of the examples include the FractalNet \cite{larsson2016fractalnet} and the ResNet \cite{he2016deep}.

Although model diversity can be increased with higher dropout rates, it can come at a cost of reduced accuracy, which is undesirable. Therefore, it seems there is a fundamental trade-off between model diversity and the accuracy of individual models in MC dropout. %~\cite{brown2005managing}. 
%In this case, the calibrated uncertainty estimates comes at the cost of predictive accuracy, which is highly undesirable. 
In practice, the dropout rate can be treated as an hyper-parameter chosen based on NLL on validation data.

\section{Experiments}\label{sec_4}
We empirically evaluate the performance of \textit{MC dropBlock}, \textit{MC dropChannel} and \textit{MC dropLayer}, and compare them to \textit{MC dropout} and \textit{deep ensembles}. Unless otherwise stated, the following experimental setup applies to all of our experiments. 

\textbf{Model} \quad In this paper, we use the PreAct-Resnet \cite{he2016identity} for all our experiments, changing only the types of dropout. 
We refer to the preAct-ResNet trained without dropout as a \textit{deterministic} model. \textit{MC dropout}, \textit{MC dropBlock} and \textit{MC dropChannel} models are implemented through inserting the corresponding dropout layers with a constant $p$ before each convolutional layer. A block size of $3 \times 3$ is used for \textit{MC dropBlock} for all the experiments. We follow \cite{ghiasi2018dropblock}
%, we use
%\begin{align}
%    \gamma = \frac{drop\_rate}{block\_size^2}\frac{feat\_size^2}{(feat\_size - block\_size +1)^2}
%\end{align}
to match up the effective dropout rate of \textit{MC dropBlock} to the desired dropout rate $p$.

\textit{MC dropLayer} is implemented through randomly dropping out entire ResNet blocks at a constant $p$, which is similar to the Stochastic Depth Net \cite{huang2016deep}. However, the Stochastic Depth Net does not perform sampling at test time. In addition, we empirically observe that, dropping out downsampling ResNet blocks during testing is harmful to the quality of uncertainty estimates. 
This is in direct agreement with \cite{veit2016residual}, in which Veit et al. demonstrated dropping out downsampling blocks significantly reduces classification accuracy\footnote{In their experiments, ResNet blocks are only dropped out during testing, but not training.}. Hence, downsampling ResNet blocks are only dropped out during training. 
As a further justification for dropping out entire blocks, \cite{veit2016residual} also showed that ResNet blocks do not strongly depend on one another, and behave like ensembles of shallower networks. Consequently, ensembles of models sampled through \textit{MC dropLayer} can achieve more diversity. 
An additional advantage is computational efficiency. For example, \textit{MC dropLayer} with $p= 0.3$ would train $30\%$ faster than that of the other methods.

For a full Bayesian treatment, we also insert a dropout layer before the fully connected layer at the end of the NNs. For all our experiments, the dropout rate of this layer is set to be $0.1$. 
To ensure a fair comparison, we also include the dropout layer before the fully connected layer of the ``deterministic'' models. 
For all models with dropout of all types, we sample $30$ times at test-time for Monte Carlo estimation. Lastly, we implement \textit{deep ensembles} by training $5$ deterministic NNs with random initializations. Although integrated into training \textit{deep ensembles} in the original paper in order to further enhance the quality of uncertainty estimates, we find that adversarial training hampers both the calibration and classification performance significantly, and thus do not incorporate it in training. This observation is in agreement with a recent finding~\cite{Tsipras2019Robustness}.

\textbf{Datasets} \quad We conduct experiments using the SVHN~\cite{netzer2011reading}, CIFAR-10 and CIFAR-100~\cite{krizhevsky2009learning} datasets with the standard train/test-set split. Validation sets of $10000$ and $5000$ samples are used for SVHN and the CIFARs respectively. To examine the performance of the proposed methods with models of different depth, we use the 18-, 50- and 152-layer PreAct-ResNet for SVHN, CIFAR-10 and CIFAR-100.

\textbf{Training} \quad  We perform preprocessing and data augmentation using per-pixel mean subtraction, horizontal random flip and $32 \times 32$ random crops after padding with 4 pixels on each side. We used stochastic gradient descent (SGD) with $0.9$ momentum, a weight decay of $10^{-4}$ and learning rate of $0.01$, and divided it by 10 after 125 and 190 epochs (250 in total) for SVHN and CIFAR-10, and after 250 and 375 (500 in total) for CIFAR-100. All the results are computed on the test set using the model at the optimal epoch based on validation accuracy. Lastly, we treat the dropout rate as a hyper-parameter and conduct a grid search with $0.05$ interval for optimal dropout rate based on NLL. 

\subsection{Evaluation of Uncertainty Estimates}
\begin{figure}
\centering
\begin{subfigure}{.3295\textwidth}
\centering
\includegraphics[width=1\linewidth]{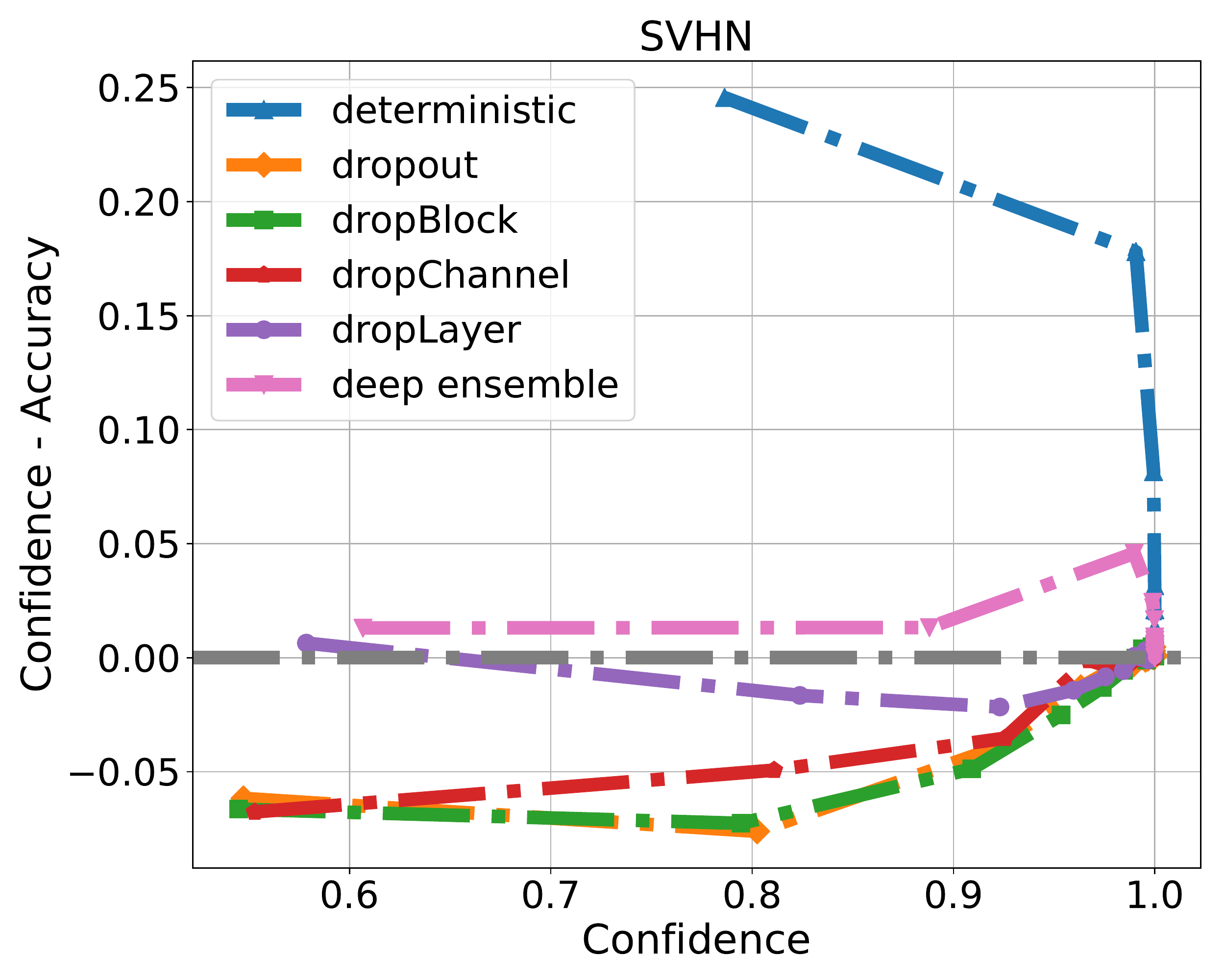}
\end{subfigure}
\begin{subfigure}{.3295\textwidth}
\centering
\includegraphics[width=1\linewidth]{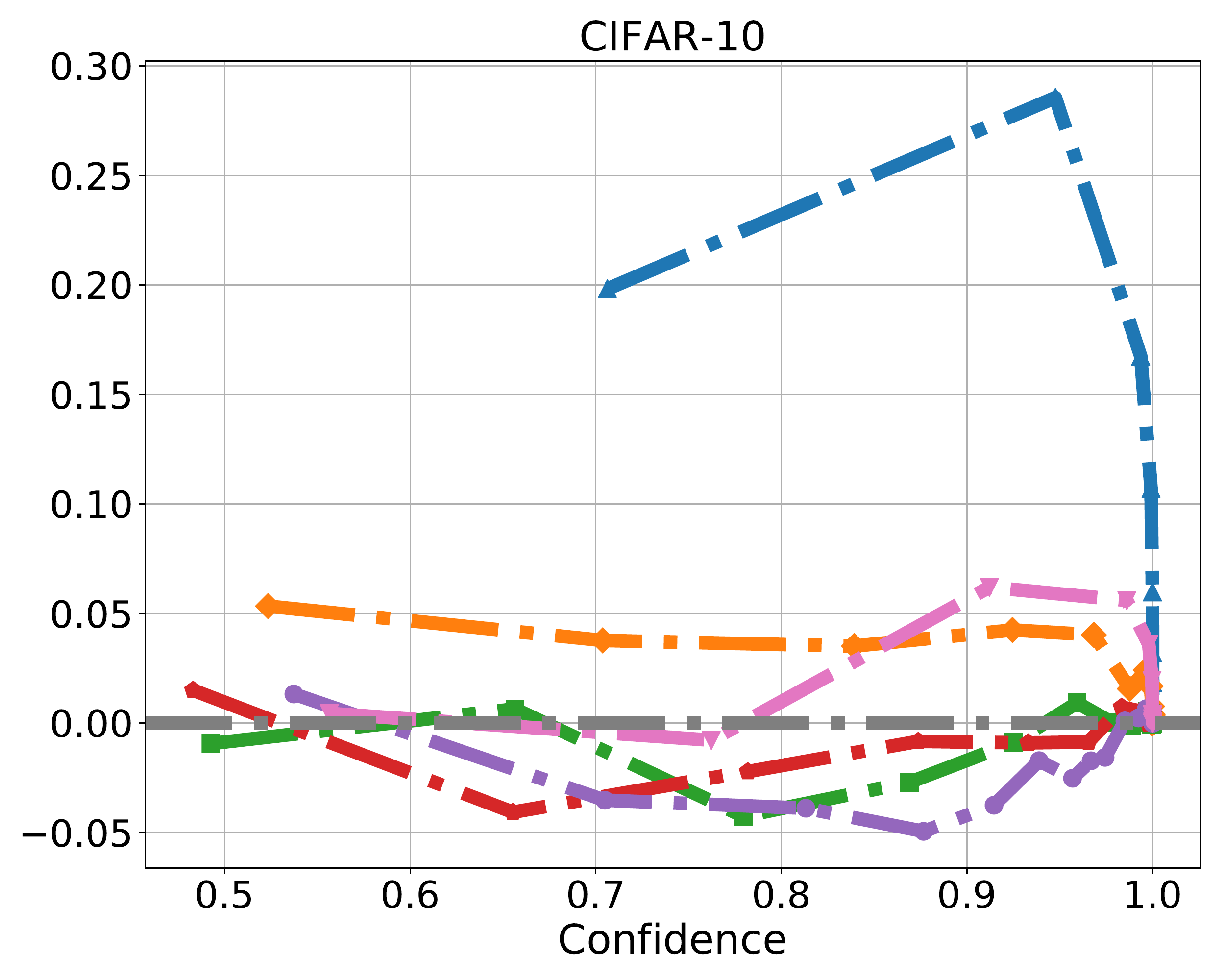}
\end{subfigure}
\begin{subfigure}{.3295\textwidth}
\centering
\includegraphics[width=1\linewidth]{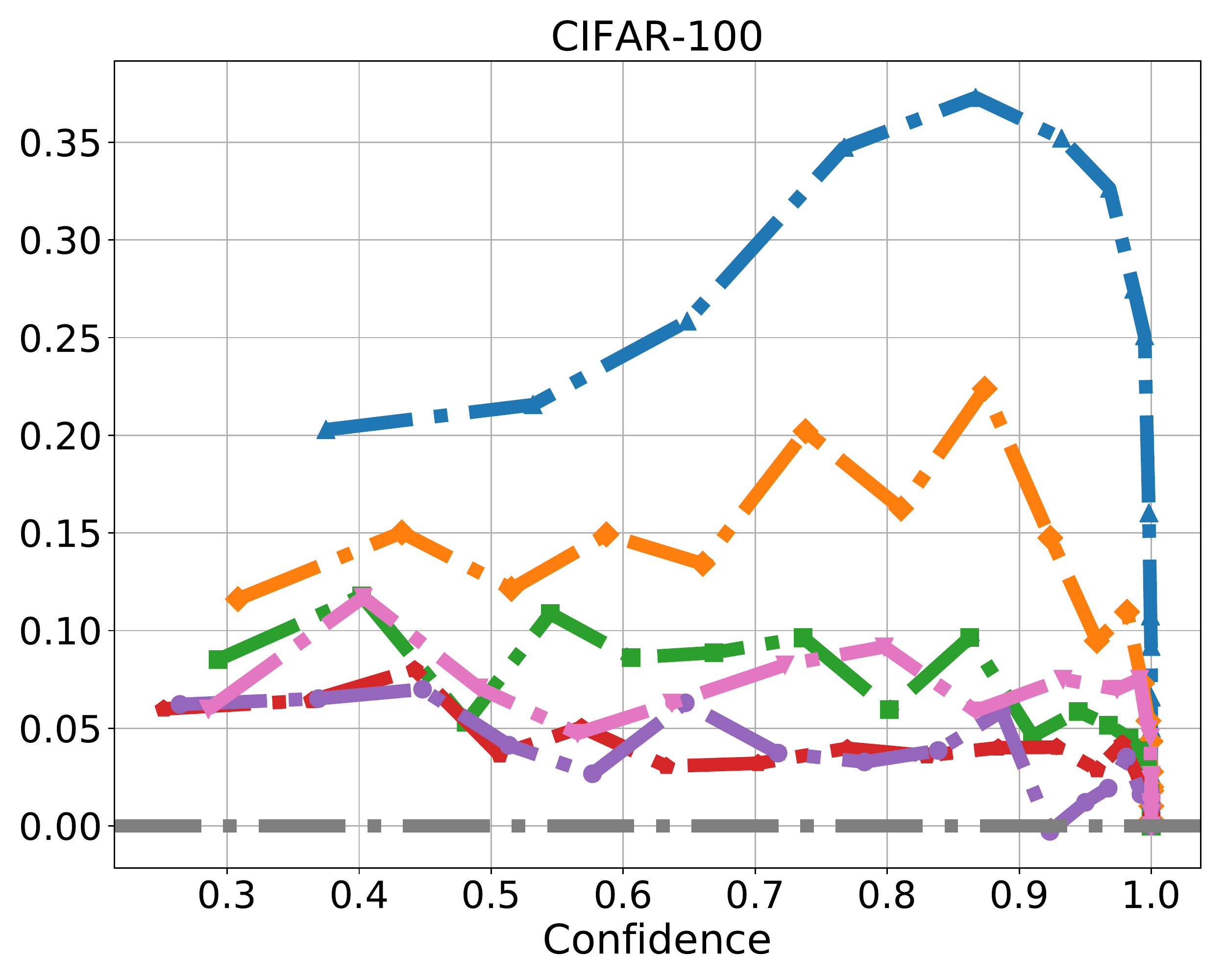}
\end{subfigure}
\caption{Reliability diagrams of predictions produced by difference models. Models with structured dropout produce better calibrated predictions, as evidenced by curves that are closer to the x-axis.}
\label{fig:ece_plot}
\end{figure}

In addition to the expected calibration error (ECE), we use the Brier score and the negative log-likelihood (NLL) to evaluate the quality of the uncertainty estimates \cite{lakshminarayanan2017simple}. Classification accuracy is also computed. Following \cite{naeini2015obtaining}, we partition predictions into 20 equally spaced bins and take a weighted average of the bins’ accuracy and confidence difference to estimate ECE. Table \ref{result_table} summarizes the quality of uncertainty estimates produced by various models. The error bars are obtained through bootstrapping the test sets. In addition, to visualize calibration performance, we show the reliability diagrams in Figure \ref{fig:ece_plot} \cite{guo2017calibration}, which are plots of the difference between accuracy and confidence against confidence. The closer the curve to the x-axis, the more calibrated the model predictions are. 

\begin{table}[]
\centering
\caption{Results on benchmark datasets comparing confidence estimates produced by different types of methods. The lower the Brier score, NLL and ECE, the better the uncertainty estimates. The top-2 performing results for each metric is bold-faced. \textit{MC dropChannel} and \textit{MC dropLayer} are the best performing methods in general. The "Dropout Rate" rows in the table corresponds to the optimal $drop\_rate$ found by grid search.}
\begin{adjustbox}{width=1\textwidth}
\begin{tabular}{c|c|ccccccc}
\hline
Datasets                  & Metric            & Deterministic     & Dropout           & DropBlock         & DropChannel       & DropLayer         & Deep Ensemble     \\ \hline \hline
\multirow{5}{*}{SVHN}     & Accuracy     & $95.7 \pm 0.1$    & $\boldsymbol{96.7 \pm 0.1}$    & $96.6 \pm 0.1$    & $\boldsymbol{96.8 \pm 0.1}$    & $96.1 \pm 0.1$    & $96.5 \pm 0.1$    \\
                          & NLL               & $0.289 \pm 0.011$ & $\boldsymbol{0.131 \pm 0.004}$ & $0.136 \pm 0.004$ & $\boldsymbol{0.128 \pm 0.004}$ & $0.147 \pm 0.004$ & $0.179 \pm 0.008$ \\
                          & Brier ($\times 10^{-3}$) & $7.41 \pm 0.21$   & $\boldsymbol{5.18 \pm 0.15}$   & $5.38 \pm 0.14$   & $\boldsymbol{5.12 \pm 0.13}$   & $5.83 \pm 0.16$   & $5.39 \pm 0.16$   \\
                          & ECE ($\times 10^{-2}$)   & $3.20 \pm 0.11$   & $1.00 \pm 0.08$   & $1.06 \pm 0.08$   & $\boldsymbol{0.86 \pm 0.08}$   & $\boldsymbol{0.53 \pm 0.07}$   & $1.09 \pm 0.08$   \\
                          & Dropout Rate      & $0.0$             & $0.35$            & $0.1$             & $0.2$             & $0.25$            & $0.0$                \\ \hline \hline
\multirow{5}{*}{CIFAR10}  & Accuracy     & $93.7 \pm 0.2$    & $93.3 \pm 0.2$    & $93.6 \pm 0.2$    & $93.6 \pm 0.2$    & $ \boldsymbol{94.1 \pm 0.2}$   & $\boldsymbol{95.2 \pm 0.2}$    \\
                          & NLL               & $0.333 \pm 0.015$ & $0.212 \pm 0.009$ & $0.198 \pm 0.007$ & $\boldsymbol{0.195 \pm 0.007}$ & $0.202 \pm 0.007$ & $\boldsymbol{0.183 \pm 0.010}$ \\
                          & Brier ($\times 10^{-3}$) & $10.6 \pm 0.40$   & $9.87 \pm 0.34$   & $9.67 \pm 0.30$   & $9.38 \pm 0.29$   & $\boldsymbol{8.96 \pm 0.30}$   & $\boldsymbol{7.52 \pm 0.29}$   \\
                          & ECE ($\times 10^{-2}$)   & $4.52 \pm 0.21$   & $1.60 \pm 0.19$   & $\boldsymbol{0.85 \pm 0.16}$ & $\boldsymbol{0.89 \pm 0.17}$   & $1.35 \pm 0.18$   & $1.47 \pm 0.16$   \\
                          & Dropout Rate      & $0.0$                & $0.2$             & $0.1$             & $0.15$            & $0.1$             & $0.0$                \\ \hline \hline
\multirow{5}{*}{CIFAR100} & Accuracy     & $74.6 \pm 0.4$    & $74.7 \pm 0.4$    & $75.4 \pm 0.4$    & $75.4 \pm 0.4$    & $\boldsymbol{76.2 \pm 0.4}$    & $\boldsymbol{78.4 \pm 0.4}$    \\
                          & NLL               & $1.42 \pm 0.03$   & $1.15 \pm 0.02$   & $1.02 \pm 0.02$   & $0.986 \pm 0.02$  & $\boldsymbol{0.975 \pm 0.02}$  & $\boldsymbol{0.910 \pm 0.020}$ \\
                          & Brier ($\times 10^{-3}$) & $4.00 \pm 0.07$   & $3.63 \pm 0.05 $  & $3.46 \pm 0.05$   & $3.40 \pm 0.05$   & $\boldsymbol{3.32 \pm 0.05}$   & $\boldsymbol{3.05 \pm 0.05}$   \\
                          & ECE ($\times 10^{-2}$)   & $15.7 \pm 0.38$   & $9.29 \pm 0.34$   & $5.45 \pm 0.35$   & $\boldsymbol{3.64 \pm 0.33}$   & $\boldsymbol{3.08 \pm 0.33}$   & $5.00 \pm 0.31$   \\
                          & Dropout Rate      & $0.0$                & $0.2$             & $0.15$            & $0.15$            & $0.25$            & $0.0$                \\ \hline 
\end{tabular}
\end{adjustbox}
\label{result_table}
\end{table}

As seen from Table \ref{result_table} and Figure \ref{fig:ece_plot}, \textit{MC dropLayer} and \textit{MC dropChannel} offer the best confidence calibration, together with good accuracy. 
%\textit{MC dropChannel} is better for shallower networks while \textit{MC dropLayer} is superior for deeper networks. 
Remarkably, both of these methods achieve even lower ECE than that of \textit{deep ensemble}, which requires $5$ times more computational resource. We also find that \textit{MC dropLayer} tends to be less sensitive to the choice of dropout rate (see Appendix A). 
While \textit{MC dropout} is able to achieve comparable performance to the proposed methods on the SVHN dataset, the confidence values produced are much worse in terms of every evaluation metric for both of the CIFAR datasets. Furthermore, comparatively, the performance of \textit{MC dropout} becomes worse as the dataset used becomes more difficult, from SVHN to CIFAR-100. Lastly, as evident from moderately increased classification accuracy over deterministic models, all types of dropout methods can be incorporated into architectures for uncertainty estimates with no accuracy penalty.

\textbf{Discussion} \quad We believe the relatively good performance of \textit{MC dropout} on SVHN compared to the CIFAR datasets is because the former task is relatively easier and thus the model can still predict accurately at an aggressive dropout rate of $0.35$ at which even regular dropout can produce acceptably diverse sampled models. 
In contrast, as observed during our experiments, while using larger dropout rates for the more difficult CIFAR datasets can lead to more calibrated predictions, classification accuracy and NLL suffer significantly (see Appendix A). 
Hence, naively increasing the dropout rate does not always lead to better performance. % since optimal dropout rate is determined based on NLL. 
Furthermore, we believe the results for \textit{MC dropBlock} can be improved by optimizing the choice of block size. 
A pre-fixed block size of $3 \times 3$ can be too small for the upstream convolutional layers where the size of feature maps are much larger than the block size, and too large for the last few downstream layers where the feature maps are comparable to the block size. 
Indeed, we observe in our experiments that when larger dropout rates are used for \textit{MC dropBlock}, classification accuracy drops drastically, indicating the possibility that too much information is discarded to make a good prediction (see Appendix A).

\subsection{Ensemble Diversity}\label{diversity}
In order to gain more insight about the improvement obtained with structured dropout, we investigate the diversity of the models sampled from dropout at all scales. 
We illustrate using models trained on CIFAR-10. For a fair comparison, we fix the dropout rate for all models to $0.1$, so that all the models trained with different types of dropout have the same complexity in terms of total effective number of parameters. 
We note however since \textit{deep ensemble} does not use dropout, it effectively has a larger capacity than all the dropout-trained models for this analysis.

\begin{figure}
\centering
\begin{subfigure}{.495\textwidth}
\centering
\includegraphics[width=1.0\linewidth]{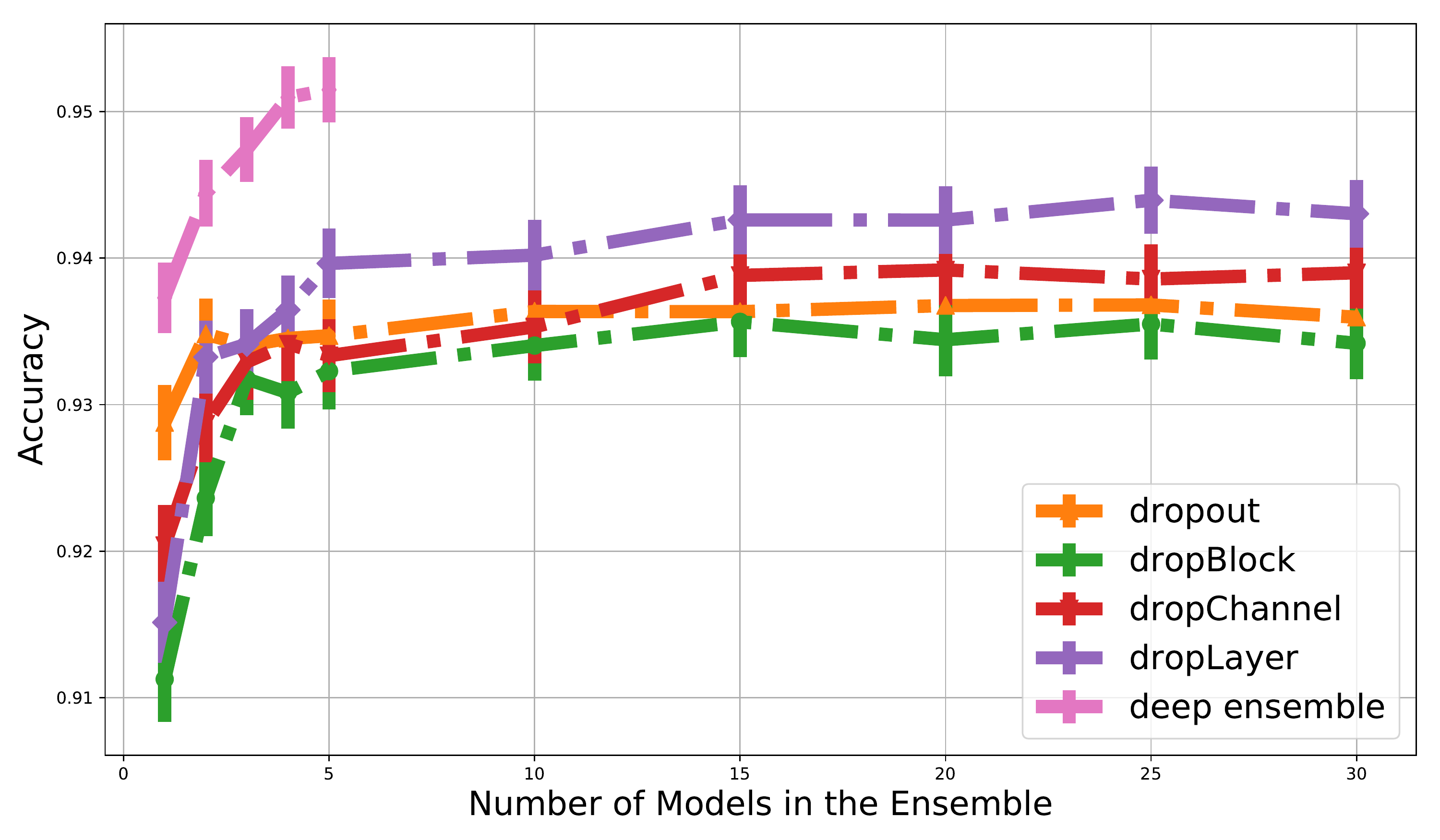}
\end{subfigure}
\begin{subfigure}{.495\textwidth}
\centering
\includegraphics[width=1.0\linewidth]{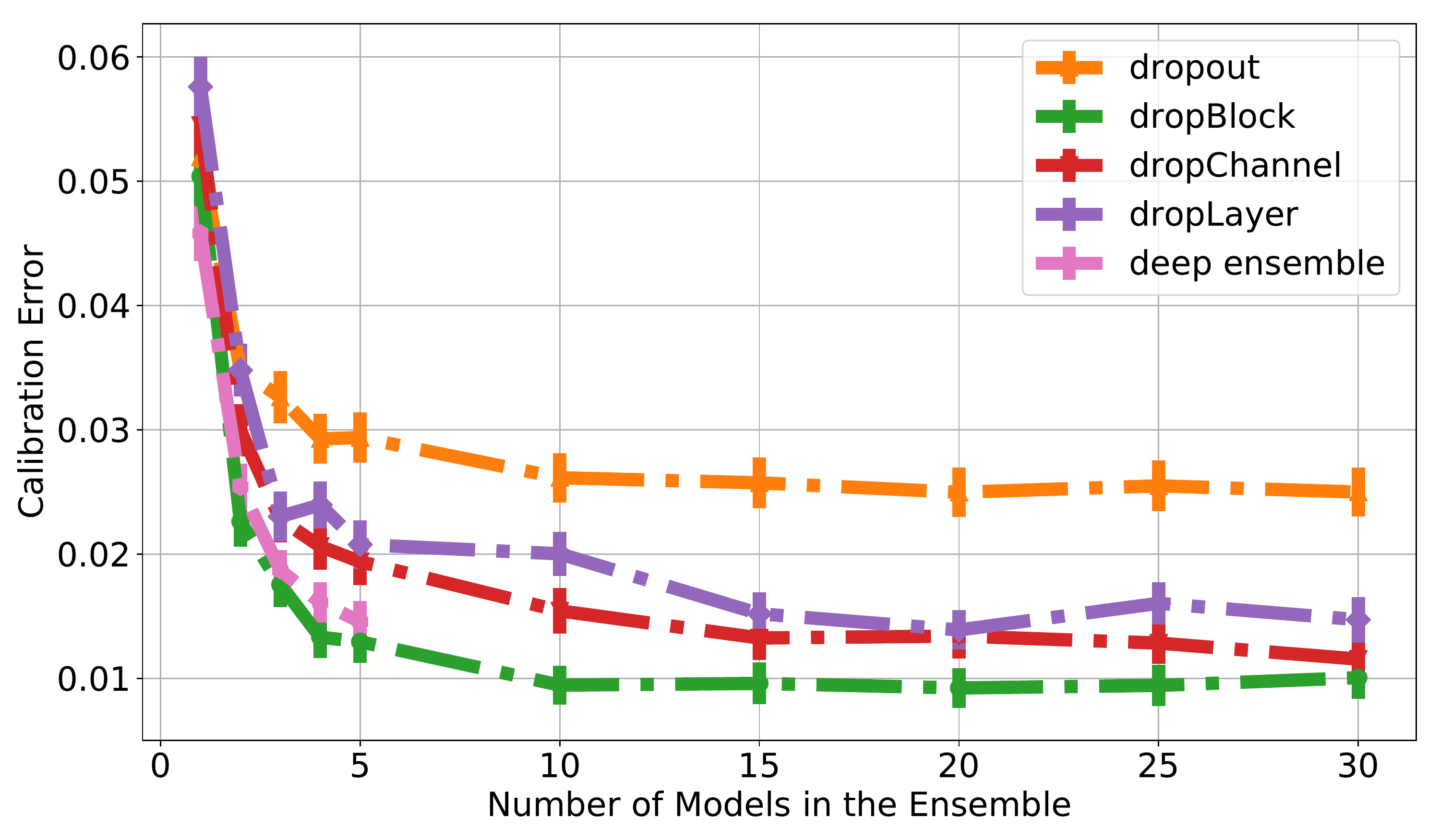}
\end{subfigure}
\caption{Test accuracy (left) and ECE (right) against number of models for ensemble prediction at test time on CIFAR-10. For varioius versions of dropout, this corresponds to the number of different MC dropout instantiations at test time of the same model. For \textit{deep ensemble}, the x-axis is the number of models trained with random initializations and SGD. The error bars are bootstrap estimated standard deviation on the testset. Models trained with structured dropout methods benefit much more from ensemble averaging.}
\label{fig:model_div}
\end{figure}

As seen from Figure \ref{fig:model_div} (\textit{Left}), the gain in accuracy by having a larger number of test-time MC samples is much smaller for regular dropout compared to that of structured dropout techniques. 
This plot also provides some insights about the average ambiguity (or model diversity) because, according to Equation \ref{eqn:decomposition}, average ambiguity is the difference between the average MSE of an individual model (which corresponds to the average accuracy of the ``one model ensemble'', the left-most points on the curves of Figure \ref{fig:model_div} (\textit{Left})) and the MSE of the ensemble of models. 
Recall that MSE is highly correlated with classification accuracy.
Thus the amount of increase in the curves of Figure \ref{fig:model_div} (\textit{Left}) are reflective of the amount of model diversity, suggesting structured dropout techniques promote more diversity than dropout.
We also see the improvement in accuracy obtained through \textit{deep ensemble}, which again is a computationally expensive strategy that relies on multiple models trained with different random initializations and SGD intantiations. 
Surprisingly, the net gain in accuracy for models obtained through structured dropout is greater than the gain offered by \textit{deep ensembling}. 
In terms of ECE, from Figure \ref{fig:model_div} (\textit{Right}), when the number of models in the ensemble is one, all approaches are very poorly calibrated. 
However, ECE decreases sharply as the number of models in the ensemble increases. 
The improvement is much less significant for \textit{MC dropout}, likely due to limited diversity.
Again, the relative improvements for models sampled with structured dropout exceeds that of the \textit{deep ensemble}. 
In fact, with five models in the ensemble, \textit{MC dropBlock} achieves a calibration error lower than \textit{deep ensemble}.
%Lastly, we observe that ECE is the lowest for \textit{deep ensemble} when number of samples is 1 because of a smaller MSE as suggested by Equation \ref{eqn:ECE}. However, the relative improvement is not as noteworthy as that achieved by having more than one model, confirming our claim in Section \ref{structure} that the enhancement of \textit{deep ensemble} over \textit{MC dropout} comes mainly from more model diversity. 

\begin{table}[]
\caption{Interrater Agreement (IA) of models with diffrent types of dropout with $0.1$ dropout rate. The lower the IA, the more diverse the predictions of the models. Dropout produces models with larger IA, hence less model diversity, than structured dropout techniques.}
\centering
\begin{adjustbox}{width=0.9\textwidth}
\begin{tabular}{l|ccccc}
\hline
Dataset & Dropout & DropBlock & DropChannel & DropLayer & Deep Ensemble \\ \hline \hline
SVHN & $0.723 \pm 0.004$  & $\boldsymbol{0.582 \pm 0.004}$ & $0.670 \pm 0.005$ & $0.746 \pm 0.008$ & $\boldsymbol{0.632}$\\ \hline
CIFAR-10 & $0.696 \pm 0.003$ & $0.568 \pm 0.005$ & $0.615 \pm 0.006$ & $\boldsymbol{0.512 \pm 0.013}$ & $\boldsymbol{0.586}$ \\ \hline
CIFAR-100 & $0.790 \pm 0.002$ & $0.708 \pm 0.003$ & $\boldsymbol{0.690 \pm 0.003}$ & $0.720 \pm 0.006$ & $\boldsymbol{0.660}$ \\ \hline
\end{tabular}
\end{adjustbox}
\label{IR_table}
\end{table}

In addition to ambiguity, there have been numerous performance measures proposed in the ensemble learning community \cite{zhou2012ensemble} to explicitly quantify the diversity of model ensembles. 
In this paper, we use the Interrater Agreement (IA) as the performance measure \cite{kuncheva2003measures}. It is defined as:
\begin{align}
\kappa = 1 - \frac{\frac{1}{T}\sum^{n}_{k=1}\rho(x_k)(T - \rho(x_k))}{n(T-1)\Bar{p}(1- \Bar{p})},    
\end{align}
where $T$ corresponds to the number of models (individual classifiers), $n$ corresponds to the total number of samples in the test set, $\rho(x_k)$ is the number of classifiers that classifies the $k$-th sample correctly, and $\Bar{p}$ is the average classification accuracy of the individual classifiers. As a measure of agreement across individual classifiers, $\kappa = 1$ if the classifiers perfectly agree on the test set. The smaller the $\kappa$, the less the individual classifiers agree, and the more the model diversity. Table \ref{IR_table} summarizes IR for sampled models trained on different datasets through dropout at different scales. We also compare the results with \textit{deep ensemble}. The number of models in the ensemble is fixed to be 5 for all approaches. 
The standard deviation estimates for the dropout methods were obtained through empirically sampling 5 model ensembles. 
The IA for \textit{MC dropout} is much higher than that of the different structured dropout techniques. 
Similar to our observations above, compared to \textit{deep ensemble}, structured dropout can yield ensembles that are as diverse as the computationally expensive method of \textit{deep ensemble}.
In fact, the IA of \textit{MC DropLayer} with CIFAR-10 is smaller than that of \textit{deep ensemble}. However, for diversity achieved by structured dropout, there is no technique that is overall superior. 
For example, \textit{MC DropLayer} achieves high IA scores on the SVHN dataset, whereas yields better diversity on CIFAR-100. 
We suspect this pattern is influenced by the choice of the neural network architecture and dataset. 
For example, for the SVHN dataset, we use a relatively shallow, 18-layer, neural network, which might constrain the achievable diversity using a specific dropout strategy. 
%, making \textit{MC DropLayer} much more deterministic in nature compared to the other models. 

\subsection{Bayesian Active Learning}
To further demonstrate the merit of the proposed use of structured dropout for confidence calibration, we consider the downstream task of Bayesian active learning on the CIFAR-10 dataset. In general, active learning involves first training on a small amount of labeled data. 
Then, an acquisition function based on the outputs of models is used to select a small subset of unlabeled data so that an oracle can provide labels for these queried data. 
Samples that a model is the least confident about are usually selected for labeling, in order to maximize the information gain. 
The model is then retrained with the additional labeled data that is provided. 
The above process can be repeated until a desired accuracy is achieved or the labeling resources are exhausted.

In our experiment, we train models with structured dropout at different scales using the identical setup as described in the beginning of this section, except that only $2000$ training samples are used initially. To match up model capacity, the dropout rate is set to $0.1$ for all methods. 
After the first iteration, we acquire $1000$ samples from a pool of "unlabeled" data, and combine the acquired samples with the original set of labeled images to retrain the models. Following Gal et al. \cite{gal2017deep}, we consider three acquision functions:  \textit{Max Entropy}, $\mathbb{H}[y|\boldsymbol{x}, \mathcal{D}_{train}] = -\sum_c p(y =c|\boldsymbol{x}, \mathcal{D}_{train})\log p(y =c|\boldsymbol{x}, \mathcal{D}_{train})$, the \textit{BALD} metric (Bayesian Active Learning by Disagreement), $\mathbb{I}[y, \boldsymbol{w}|\boldsymbol{x}, \mathcal{D}_{train}] = \mathbb{H}[y|\boldsymbol{x}, \mathcal{D}_{train}] - \mathbb{E}_{p(\boldsymbol{w}|\mathcal{D}_{train})[\mathbb{H}[y|\boldsymbol{x},\boldsymbol{w}]]}$, and the \textit{Variation Ratios} metric, $\text{variation-ratio}[\boldsymbol{x}] = 1- \max_y p(y, \boldsymbol{x}, \mathcal{D}_{train})$. We repeat the acquisition process 9 times so that in the last iteration, the training set contains $10000$ images. Test accuracies are calculated before each acquisition phase. To mimic a real world scenario in which number of labeled samples is small, we do not use a validation set, and the accuracies reported for this experiment correspond to the last-epoch accuracies. Due to the small training set and hence larger variance in results, we repeat the experiment 5 times and compute the mean accuracies.

\begin{figure}
\centering
\begin{subfigure}{.495\textwidth}
\centering
\includegraphics[width=1\linewidth]{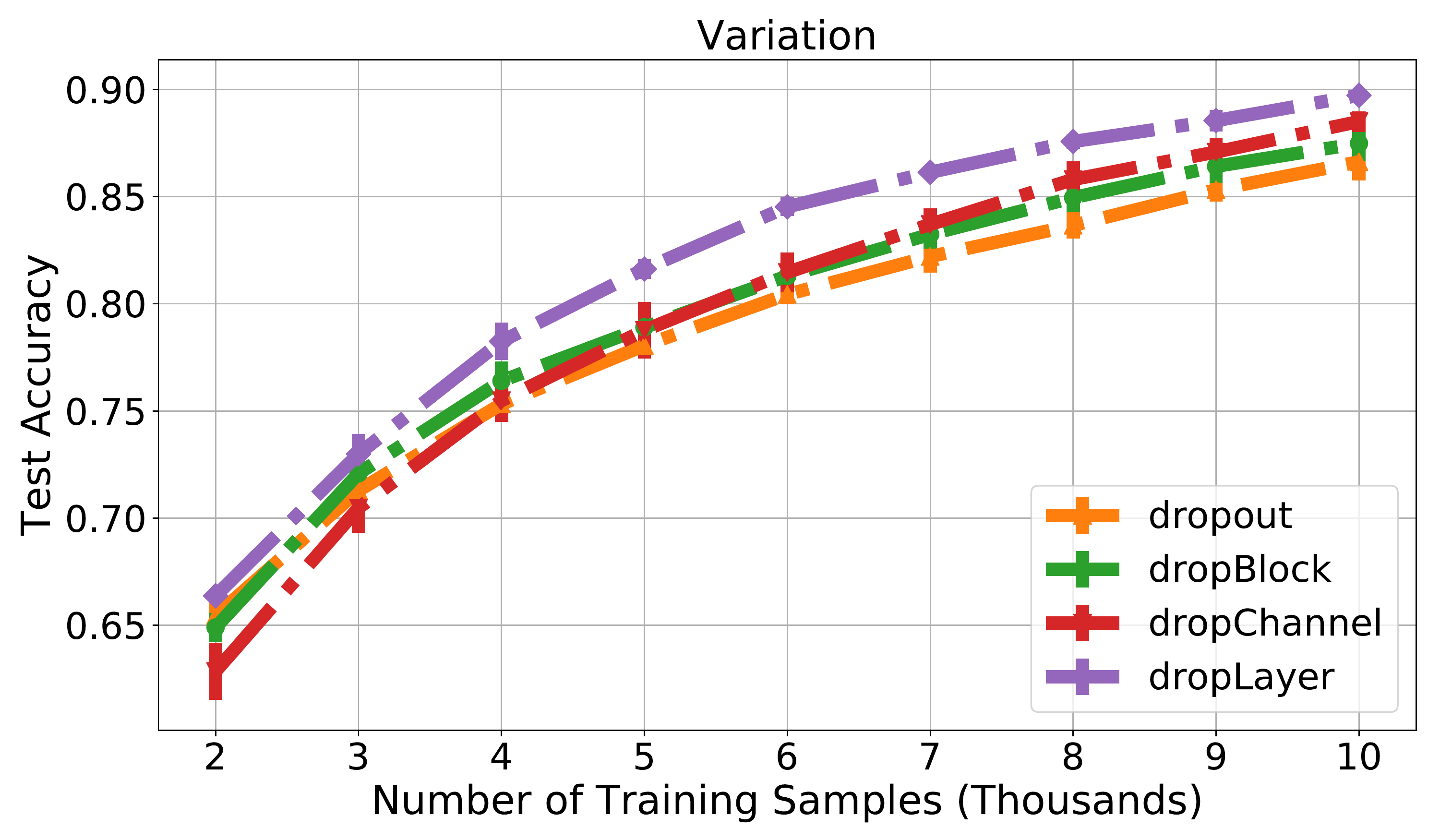}
\end{subfigure}
\begin{subfigure}{.495\textwidth}
\centering
\includegraphics[width=1\linewidth]{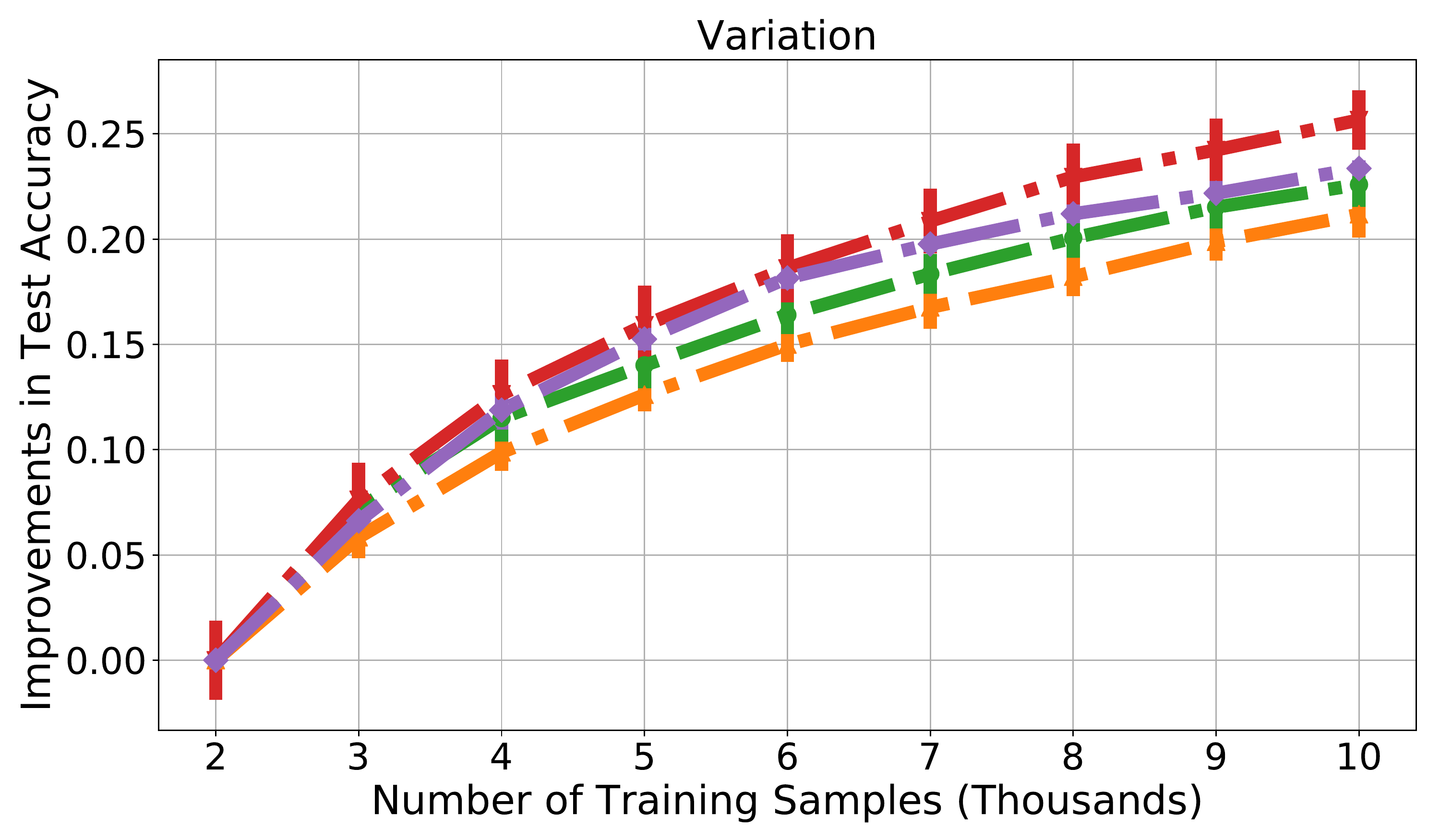}
\end{subfigure}
\caption{\textit{Left}: Test accuracy against number of training samples for models with different methods of dropout and Variation Ratios as the acquisition function on CIFAR-10. \textit{Right}: Relative improvements in test accuracy over that of the first iteration with different methods of dropout. \textit{MC dropout} yields the least improvements of all the methods.}
\label{fig:al_acc}
\end{figure}

Figure \ref{fig:al_acc} shows the test accuracy against number of training samples for different models, with Variation Ratios used as the acquisition function. It can be seen that, after the first iteration when all $2000$ training images are randomly selected, the test accuracy using \textit{MC dropout} is on par with that of other methods. 
%The accuracy of \textit{MC dropLayer} is slightly higher and that of \textit{MC dropChannel} is moderately lower. 
However, as more labeled data are added, the relative increase in accuracy is more significant for models using structured dropout compared to that of using regular dropout. 
For instance, the ulitmate increase in accuracy with \textit{MC dropout} is $21.1 \pm 1.1 \%$ compared to that of $25.7 \pm 1.4 \%$ when using \textit{MC dropChannel}. 
This suggests that the uncertainty estimates obtained with structured dropout are more useful for assessing "what the model doesn't know", thereby allowing for the selection of samples to be labeled in a way that better helps improve performance. 
%Among all models, the relative improvement for \textit{MC dropChannel} is the most significant, while \textit{MC dropLayer} performs the best. 
Similar trends are observed for the other acquisition functions (see Appendix B). Of the three acquisition functions, \textit{Max Entropy} and \textit{Variation Ratios} seem to be more effective in our experiment. 

\section{Conclusion and Future Work}\label{sec_5}
We reinterpret \textit{MC dropout} as ensemble averaging strategy, and attribute the calibration error in confidence estimates to a lack of diversity of sampled models. 
As we demonstrate in our experiments, structured dropout, which is simple-to-implement and computationally efficient, can promote model diversity and improve calibration. 
The gain in performance, however, depends on the architecture, the task, the dropout rate and the structured dropout strategy.

We are interested in further exploring several directions. Firstly, there are many other architectures for implementing \textit{MC dropLayer} \cite{larsson2016fractalnet, xie2017aggregated}, and we have only considered ResNet. 
Moreover, we used a constant dropout rate in our experiments, even though one can vary dropout rates across NNs \cite{huang2016deep} or incorporate dropout rate scheduling \cite{ghiasi2018dropblock}. How this impacts calibration is an open question. Lastly, we have only considered structured dropout in the context of CNNs, with application to computer vision tasks. We believe this idea can be extended beyond CNNs.

% This work was supported by NIH R01 grants (R01LM012719 and R01AG053949), the NSF NeuroNex grant 1707312, and NSF CAREER grant (1748377).

\bibliography{reference}
\bibliographystyle{plain}

\newpage
\section*{Appendix A: Performance of Uncertainty Estimates Against Dropout Rate}

\begin{figure}[!htb]
\centering
\begin{subfigure}{.45\textwidth}
\centering
\includegraphics[width=1\linewidth]{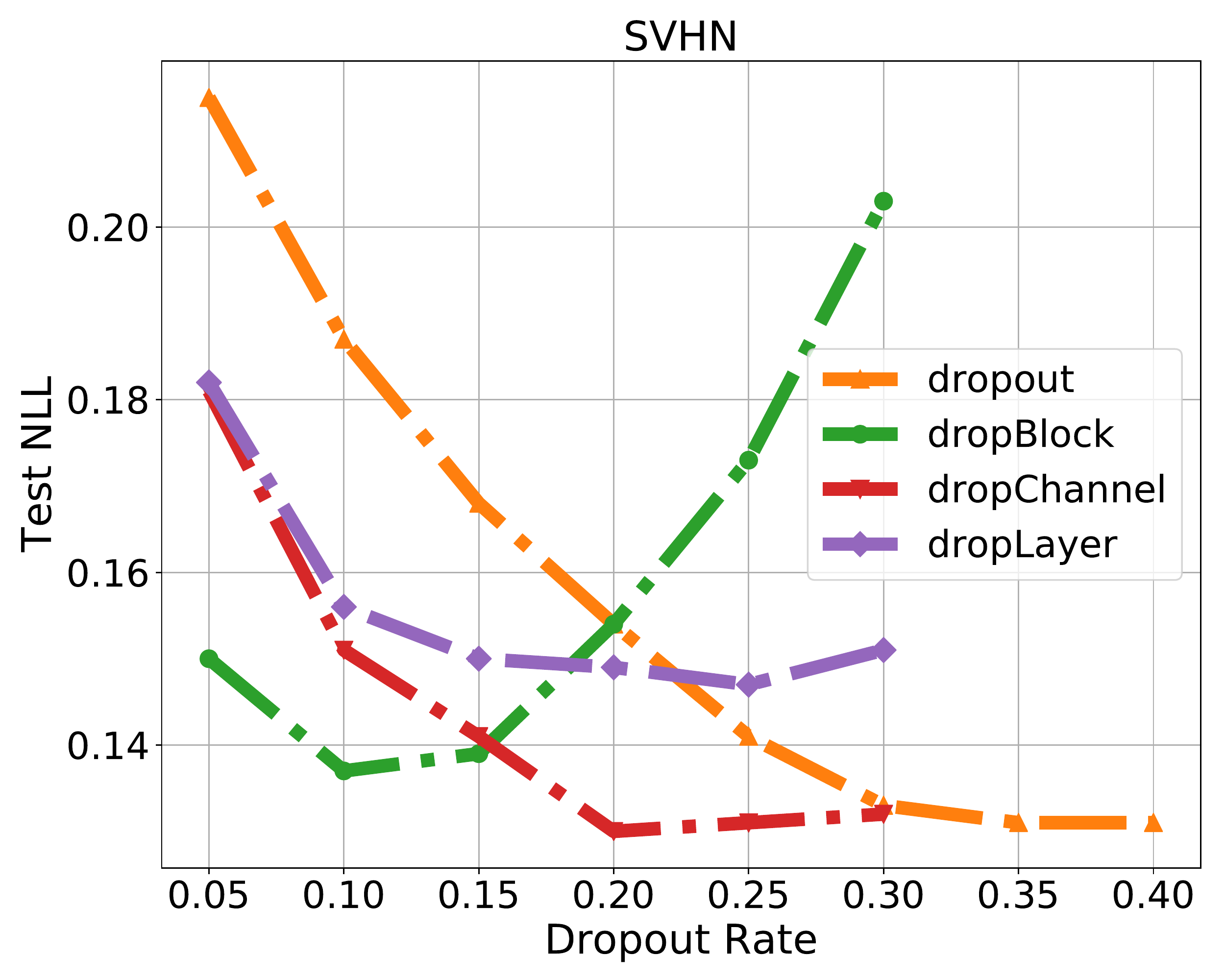}
\end{subfigure}
\begin{subfigure}{.45\textwidth}
\centering
\includegraphics[width=1\linewidth]{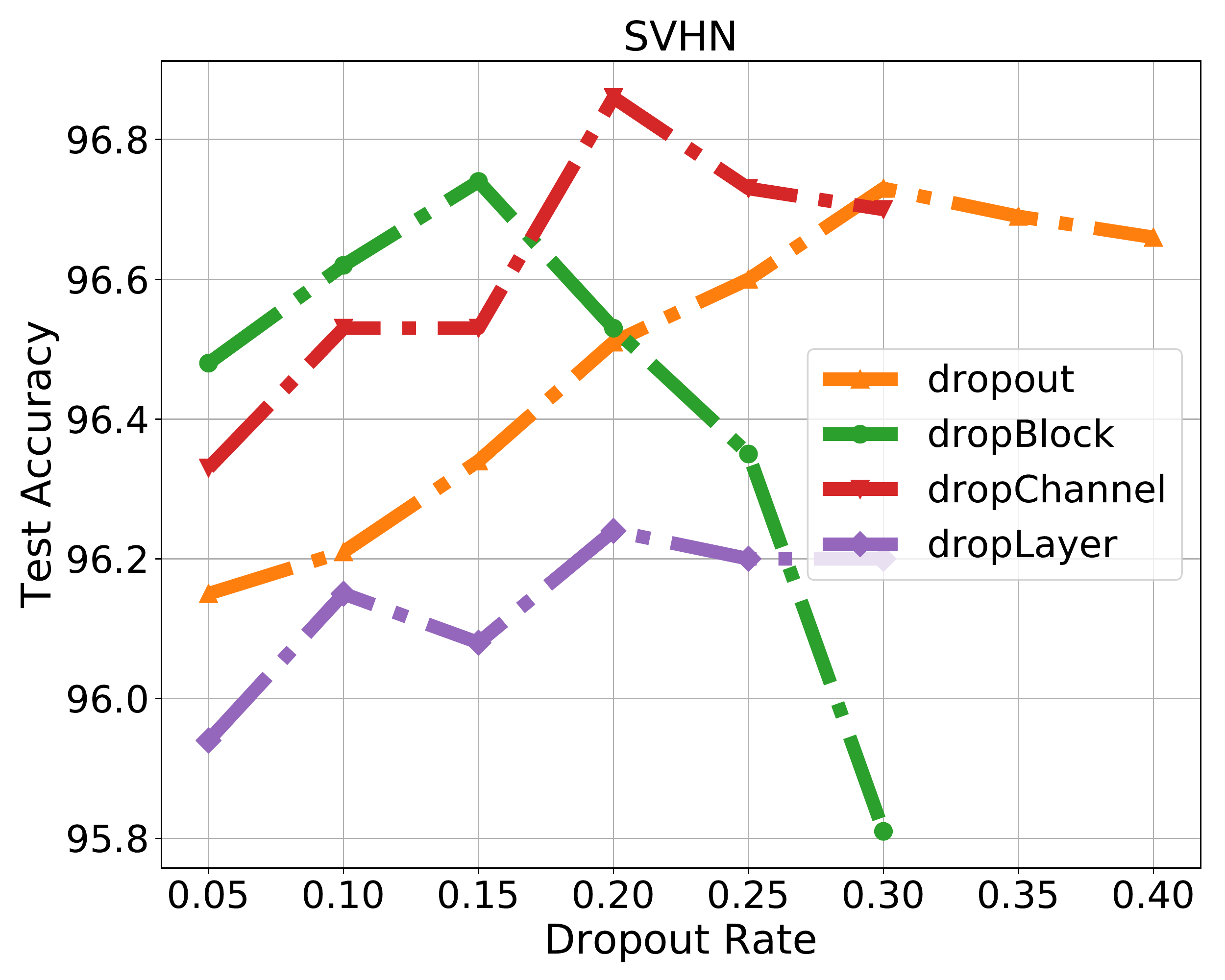}
\end{subfigure}
\begin{subfigure}{.45\textwidth}
\centering
\includegraphics[width=1\linewidth]{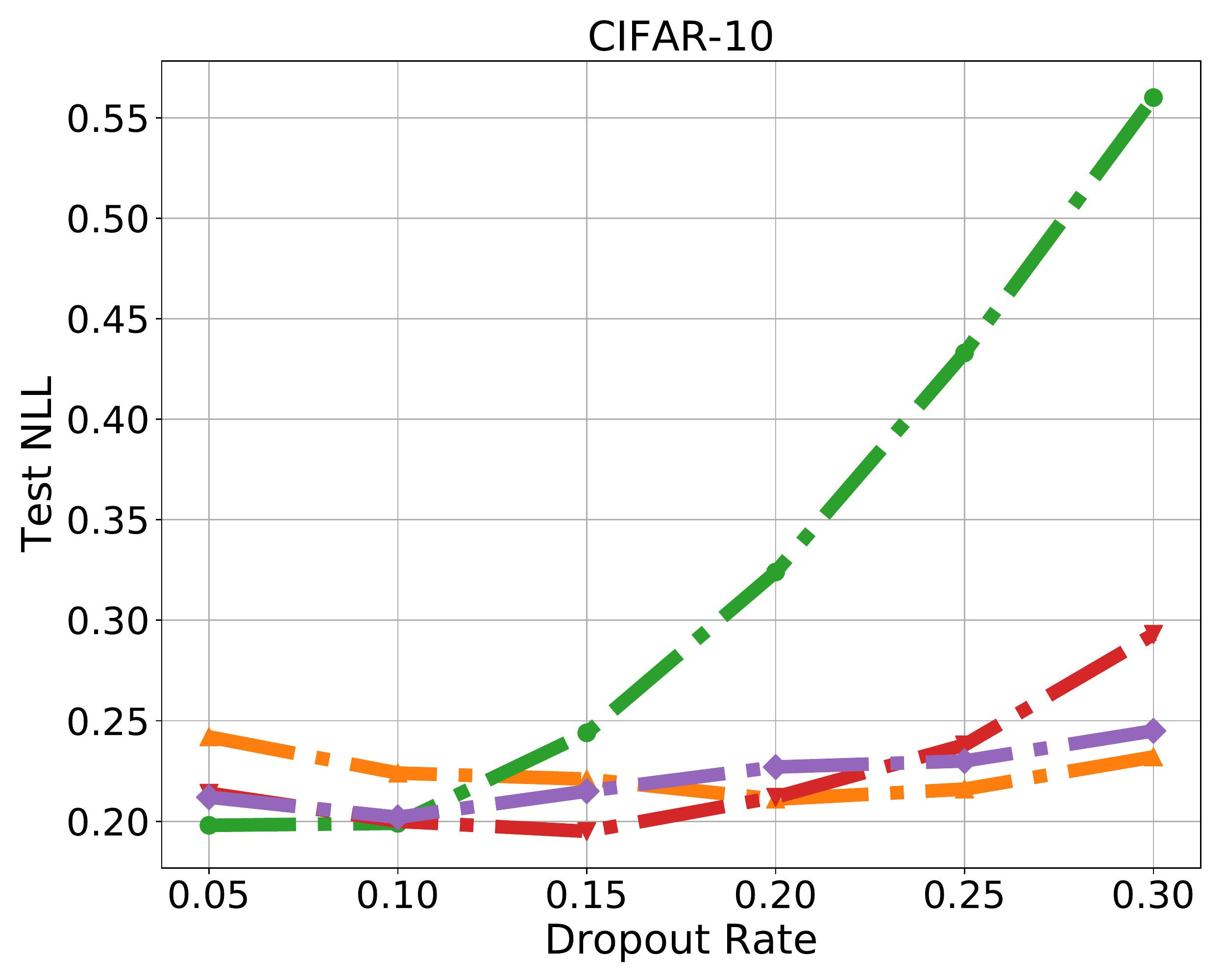}
\end{subfigure}
\begin{subfigure}{.45\textwidth}
\centering
\includegraphics[width=1\linewidth]{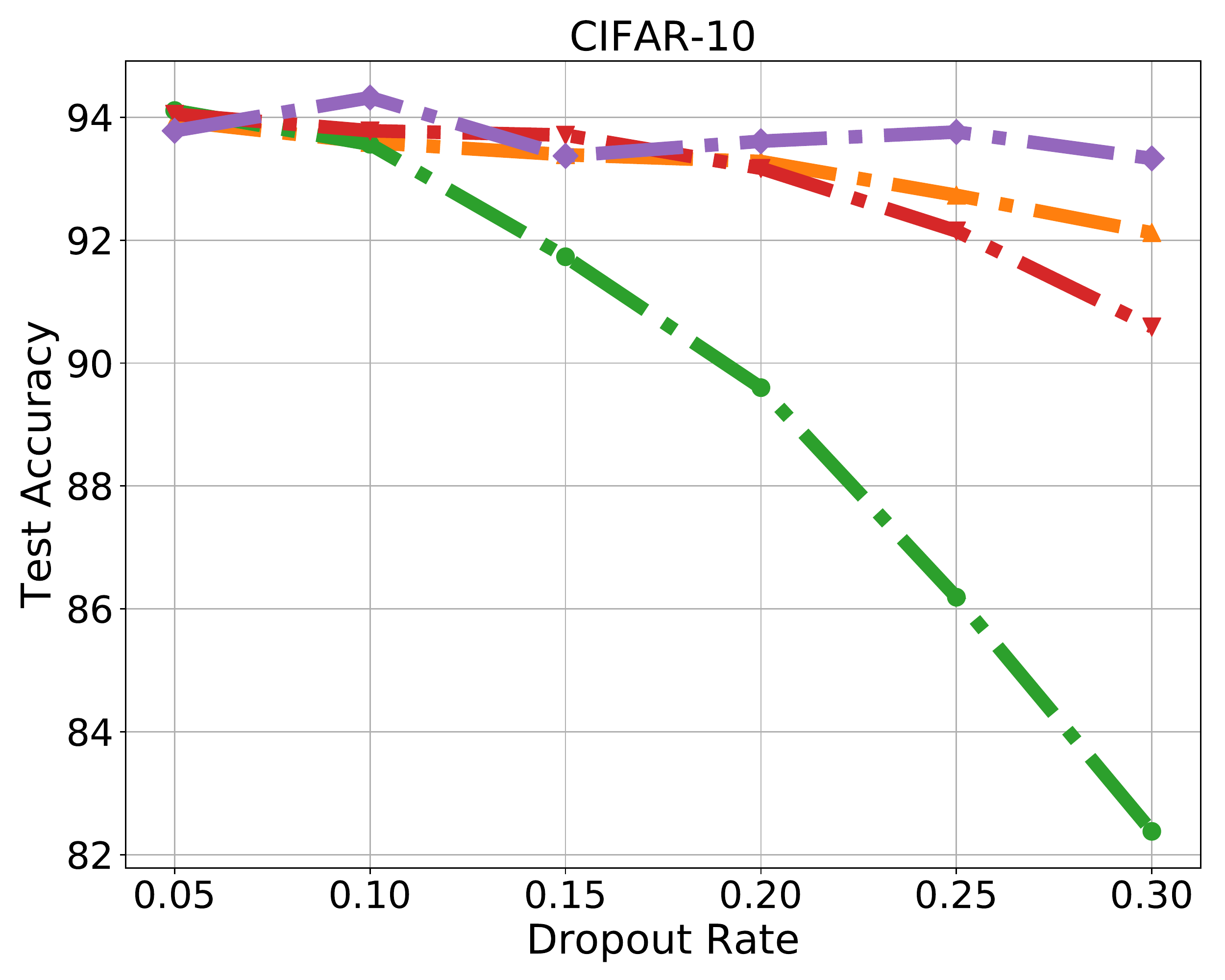}
\end{subfigure}
\begin{subfigure}{.45\textwidth}
\centering
\includegraphics[width=1\linewidth]{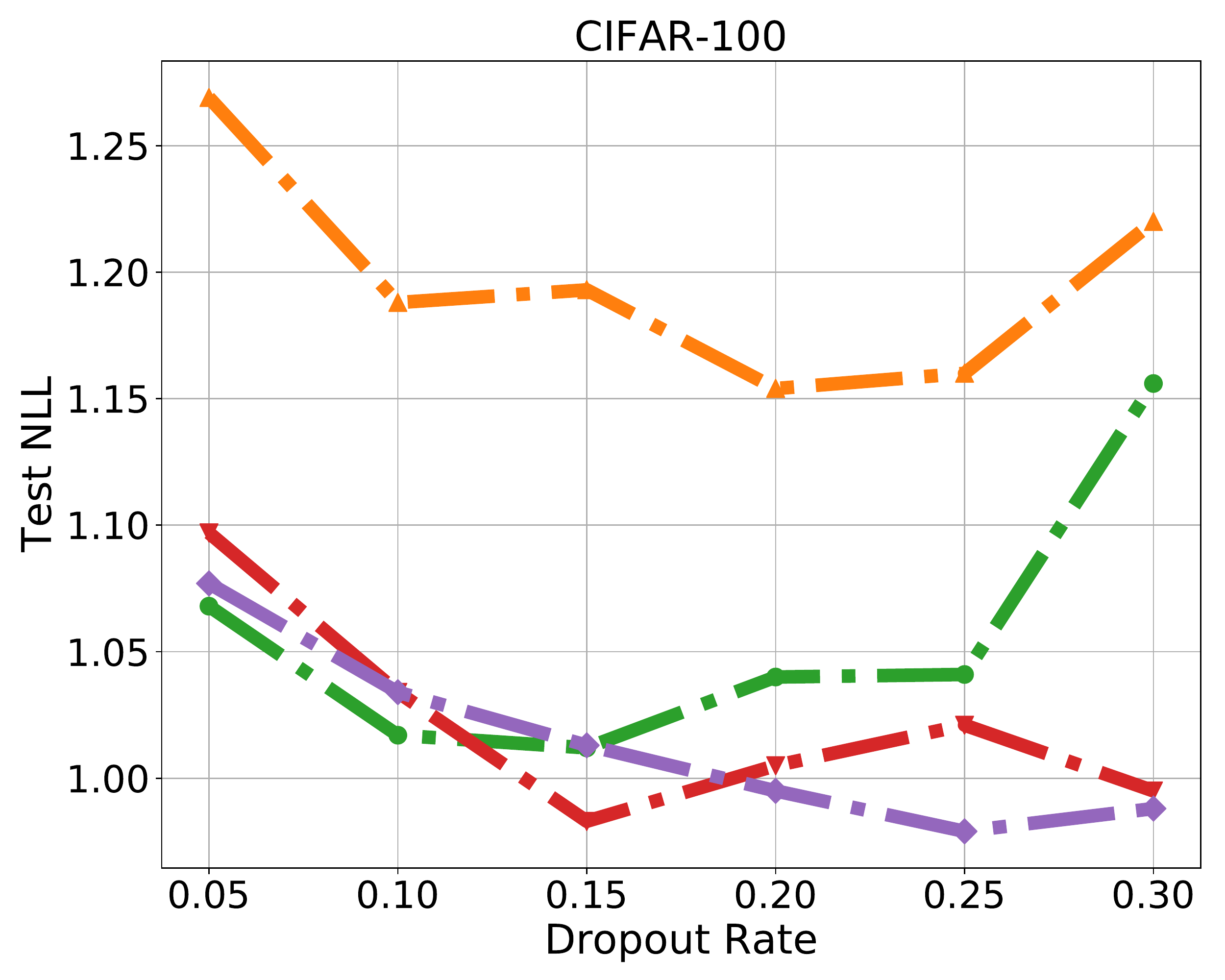}
\end{subfigure}
\begin{subfigure}{.45\textwidth}
\centering
\includegraphics[width=1\linewidth]{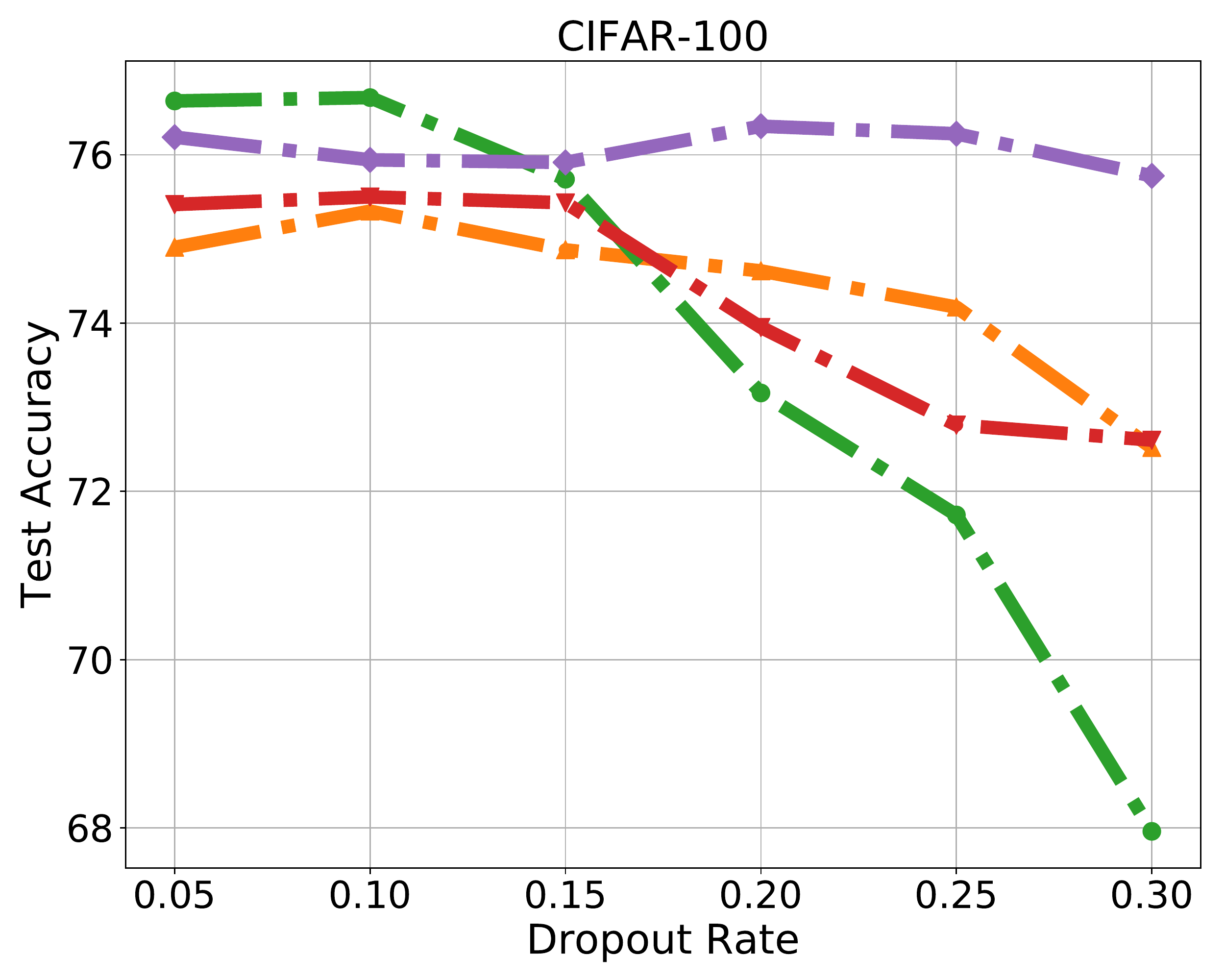}
\end{subfigure}
\caption{Plots of test time NLL (\textit{Left}) and accuracy (\textit{Right}) against dropout rate for models trained with different types of dropout on the SVHN, CIFAR-10 and CIFAR-100 datasets. 
Models trained with structured dropout can achieve better NLL performance, particularly for moderate values of the dropout rate. \textit{DropLayer} is the least sensitive to the choice of dropout rate with respect to NLL. Interestingly, the NLL drastically increases after minima on all three datasets for \textit{dropBlock}, suggesting the possibility that the block size for \textit{dropBlock} may be too large towards later convolutional layers when the size of feature maps are comparable to that of block size. }
\label{fig:nll_droprate}
\end{figure}

\newpage
\section*{Appendix B: Additional Results with Active Learning}

\begin{figure}[!htb]
\centering
\begin{subfigure}{.495\textwidth}
\centering
\includegraphics[width=1\linewidth]{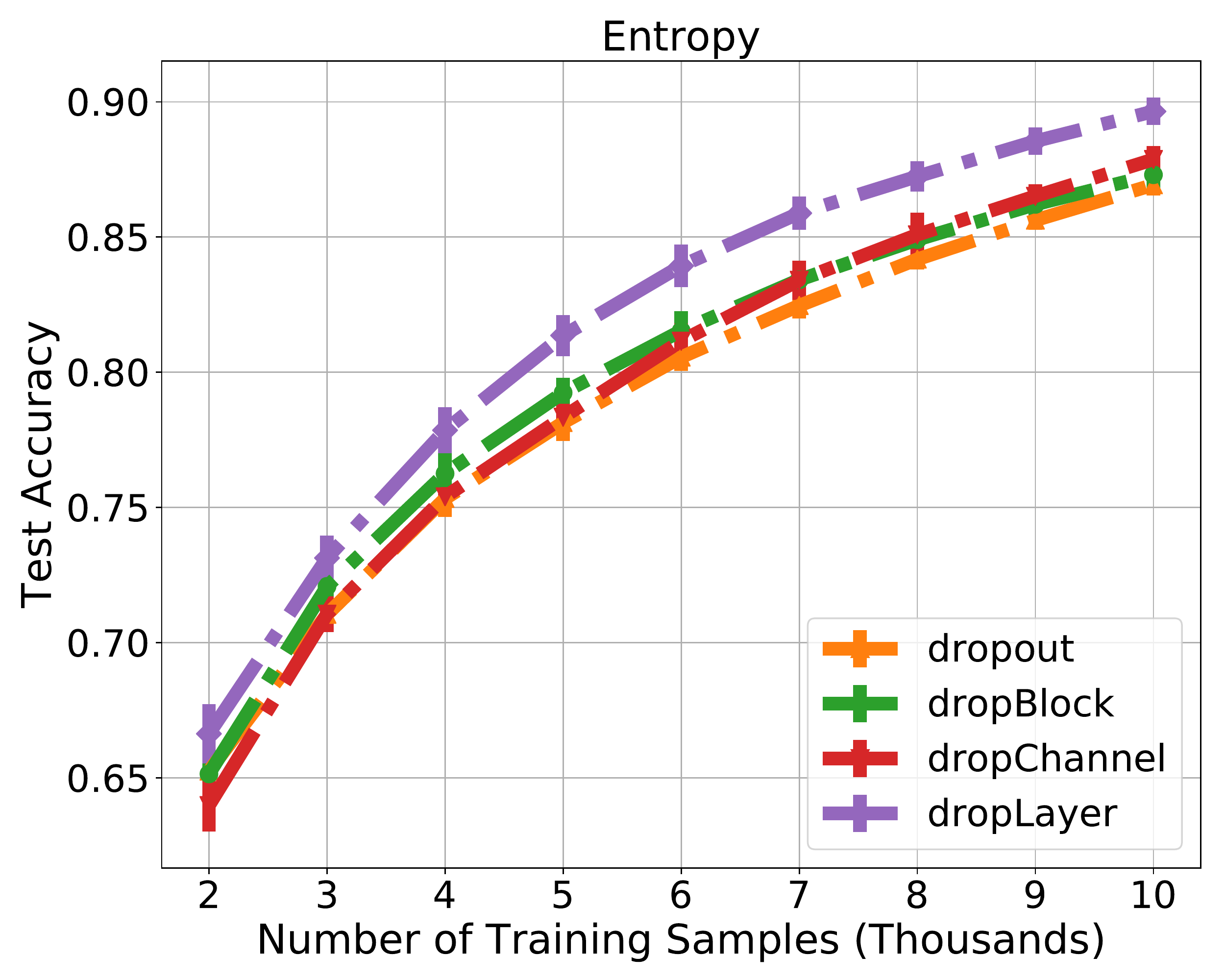}
\end{subfigure}
\begin{subfigure}{.495\textwidth}
\centering
\includegraphics[width=1\linewidth]{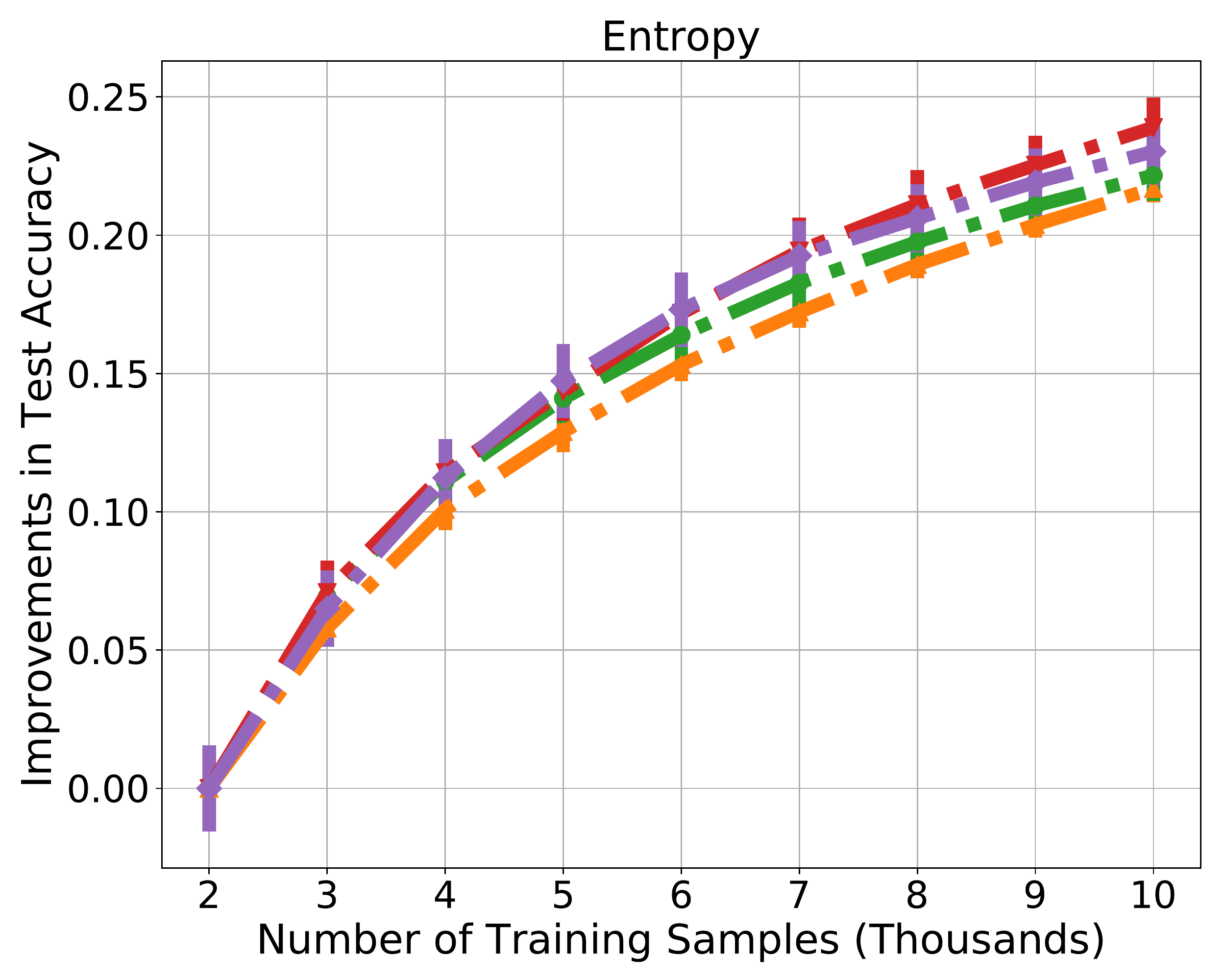}
\end{subfigure}
\begin{subfigure}{.495\textwidth}
\centering
\includegraphics[width=1\linewidth]{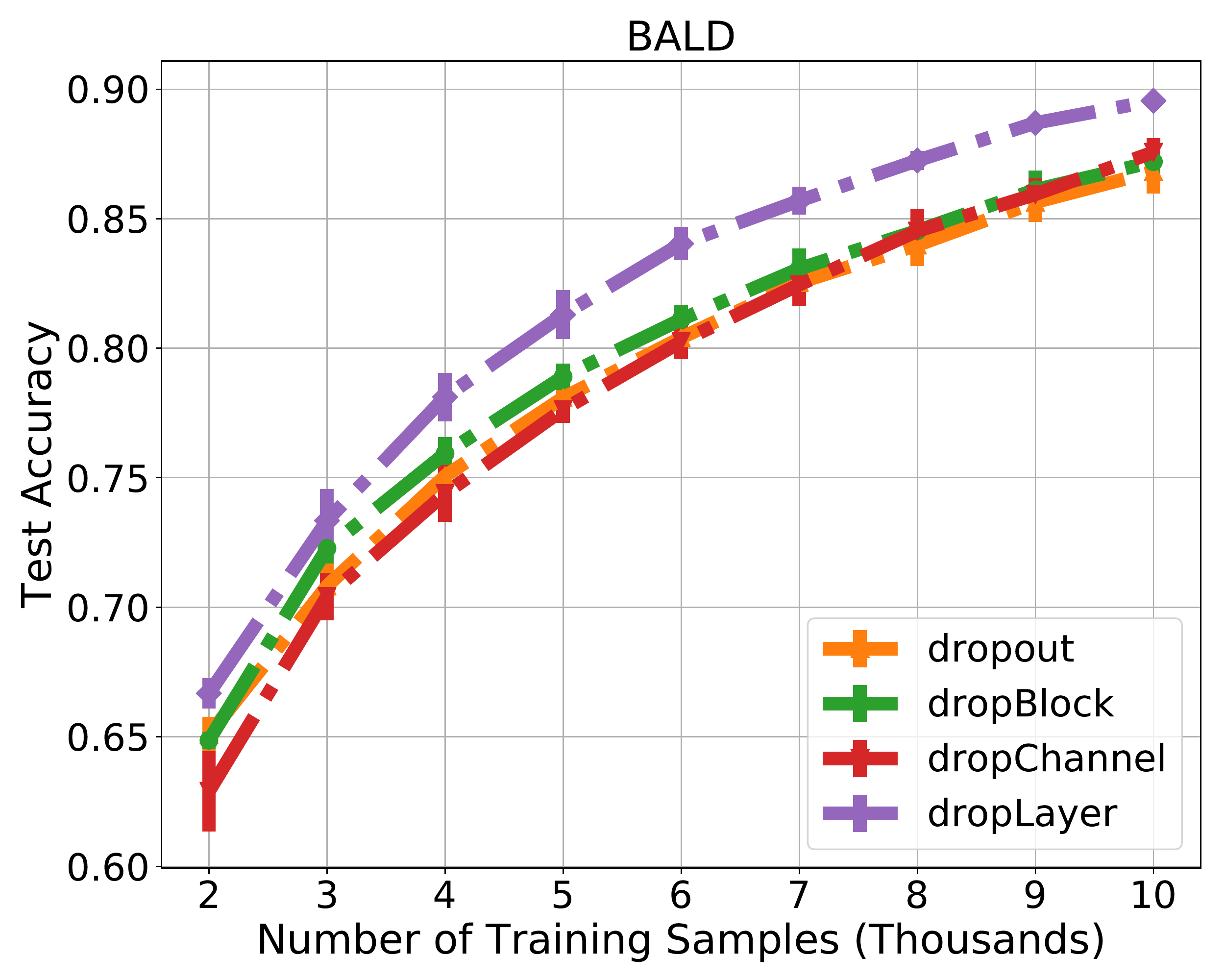}
\end{subfigure}
\begin{subfigure}{.495\textwidth}
\centering
\includegraphics[width=1\linewidth]{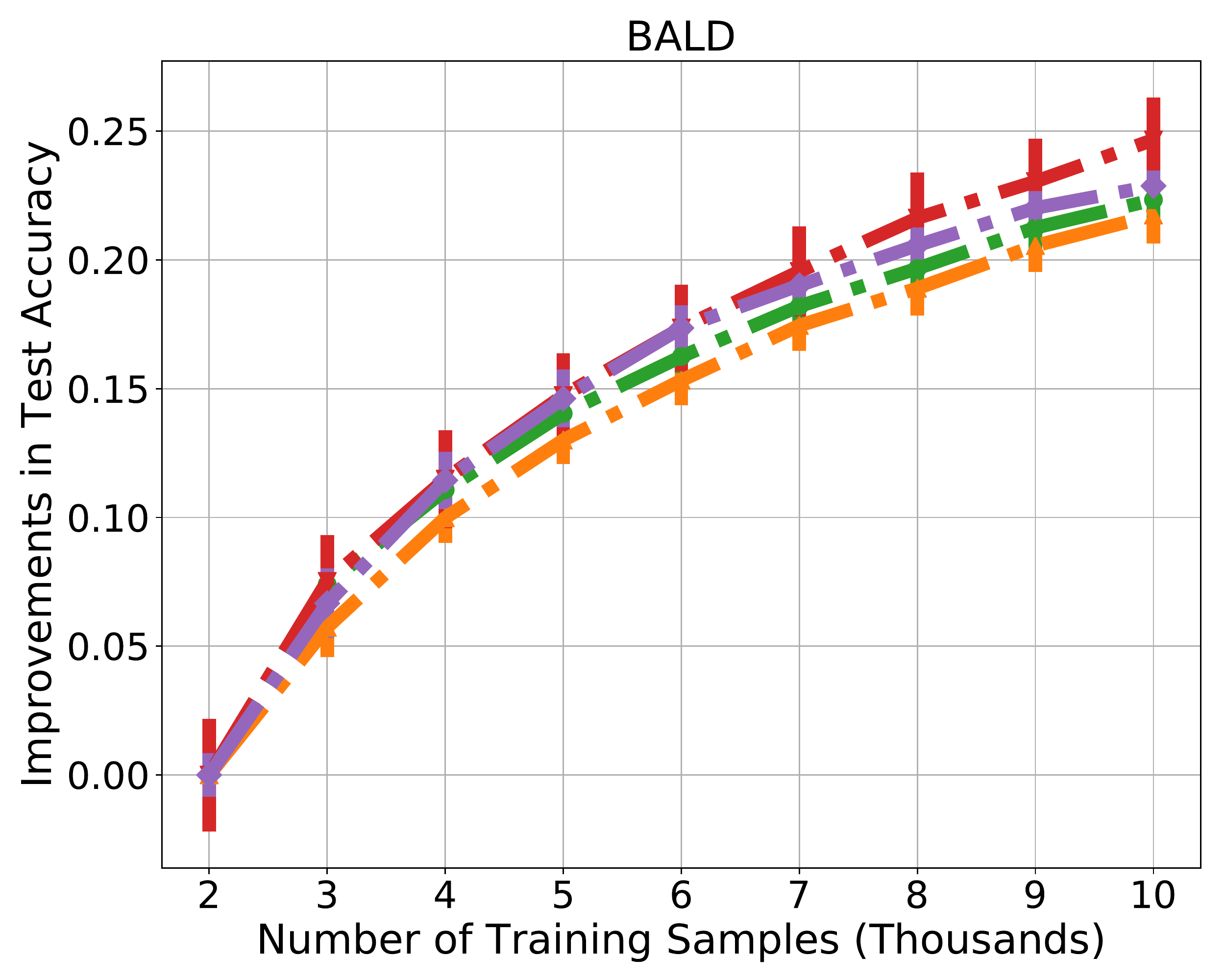}
\end{subfigure}
\caption{\textit{Left}: Test accuracy against number of training samples for models with different methods of dropout and \textit{Max Entropy} (Above) / \textit{BALD} (Below) as the acquisition function on CIFAR-10. \textit{Right}: Relative improvements in test accuracy over that of the first iteration with different methods of dropout. Similar to results obtained with Variation Ratios, \textit{MC dropout} yields the least improvements of all the methods.}
\label{fig:al_acc_more}
\end{figure}

\end{document}